\documentclass{article}

\usepackage{microtype}
\usepackage{graphicx}
\usepackage{subcaption} 
\usepackage{booktabs} 

\usepackage{hyperref}

\usepackage{array}
\usepackage{colortbl}

\usepackage[accepted]{icml2024}

\usepackage{amsmath}
\usepackage{amssymb}
\usepackage{mathtools}
\usepackage{amsthm}

\usepackage{amsfonts}
\usepackage{enumitem}
\usepackage{multirow}
\usepackage{tcolorbox}

\newcommand{\unet}[0]{\textit{U-Net}}
\newcommand{\knn}[0]{\textit{KNN}}

\newcommand{\gnn} [0]{\textit{GNN}}
\newcommand{\ens}[0]{\textit{Deep Ensembles}}
\newcommand{\mcdp}[0]{\textit{MC-Dropout}}
\newcommand{\ours}[0]{\textit{Ours}}

\definecolor{best}{RGB}{200, 200, 255}
\definecolor{secondbest}{RGB}{230, 230, 255}

\newcommand{\parag}[1]{\vspace{-3mm}\paragraph{#1}}

\newif\ifdraft
\draftfalse

\ifdraft
\newcommand{\PF}[1]{{\color{red}{\bf PF: #1}}}

\newcommand{\HL}[1]{{\color{orange}{\bf HL: #1}}}
\newcommand{\hl}[1]{{\color{orange} #1}}
\newcommand{\AL}[1]{{\color{blue}{\bf AL: #1}}}

\newcommand{\ND}[1]{{\color{blue}{\bf ND: #1}}}
\newcommand{\nd}[1]{{\color{blue} #1}}
\newcommand{\JD}[1]{{\color{green}{\bf JD: #1}}}

\else
\newcommand{\PF}[1]{}

\newcommand{\HL}[1]{}
\newcommand{\hl}[1]{#1}
\newcommand{\AL}[1]{}

\newcommand{\ND}[1]{}
\newcommand{\nd}[1]{#1}
\newcommand{\JD}[1]{}

\fi

\newcommand{\bY}{\mathbf{Y}}

\newcommand{\bx}{\mathbf{x}}

\newcommand{\by}{\mathbf{y}}
\newcommand{\bz}{\mathbf{z}}

\definecolor{bleudefrance}{RGB}{49,140,231}

\newcommand{\graybold}[1]{\textcolor{gray}{\textbf{#1}}}

\usepackage[capitalize,noabbrev]{cleveref}

\theoremstyle{plain}

\theoremstyle{definition}

\theoremstyle{remark}

\usepackage[textsize=tiny]{todonotes}

\icmltitlerunning{Enabling Uncertainty Estimation in Iterative Neural Networks}

\begin{document}

\twocolumn[
\icmltitle{Enabling Uncertainty Estimation in Iterative Neural Networks}



\icmlsetsymbol{equal}{*}

\begin{icmlauthorlist}
\icmlauthor{Nikita Durasov}{epfl}
\icmlauthor{Doruk Oner}{epfl}
\icmlauthor{Jonathan Donier}{comp}
\icmlauthor{Hieu Le}{epfl}
\icmlauthor{Pascal Fua}{epfl}
\end{icmlauthorlist}

\icmlaffiliation{epfl}{Computer Vision Laboratory, École Polytechnique Fédérale de Lausanne, Lausanne, Switzerland}
\icmlaffiliation{comp}{Neural Concept SA, Lausanne, Switzerland}

\icmlcorrespondingauthor{Nikita Durasov}{yassnda@gmail.com}

\icmlkeywords{Machine Learning, ICML}

\vskip 0.3in
]



\printAffiliationsAndNotice{This work was supported in part by the Swiss National Science Foundation.}  


\begin{abstract}

Turning pass-through network architectures into iterative ones, which use their own output as input, is a well-known approach for boosting performance. In this paper, we argue that such architectures offer an additional benefit: The convergence rate of their successive outputs is highly correlated with the accuracy of the value to which they converge. Thus, we can use the convergence rate as a useful proxy for uncertainty. This results in an approach to uncertainty estimation that provides state-of-the-art estimates at a much lower computational cost than techniques like Ensembles, and without requiring any modifications to the original iterative model.
We demonstrate its practical value by embedding it in two application domains: road detection in aerial images and the estimation of aerodynamic properties of 2D and 3D shapes.
\vspace{-2mm}
\begin{center}
\ \  \ \ \href{https://icml.cc/virtual/2024/poster/34213}{\textcolor{bleudefrance}{\texttt{poster}}}\ \ /
\ \ \href{https://github.com/cvlab-epfl/iter_unc}{\textcolor{bleudefrance}{\texttt{code}}}\ \ /
\ \ \href{https://www.norange.io/projects/unc_iter/}{\textcolor{bleudefrance}{\texttt{web}}}
\end{center}
\end{abstract}
\vspace{-4mm}

\section{Introduction}


\begin{figure}[htb]
    \includegraphics[width=0.48\textwidth]{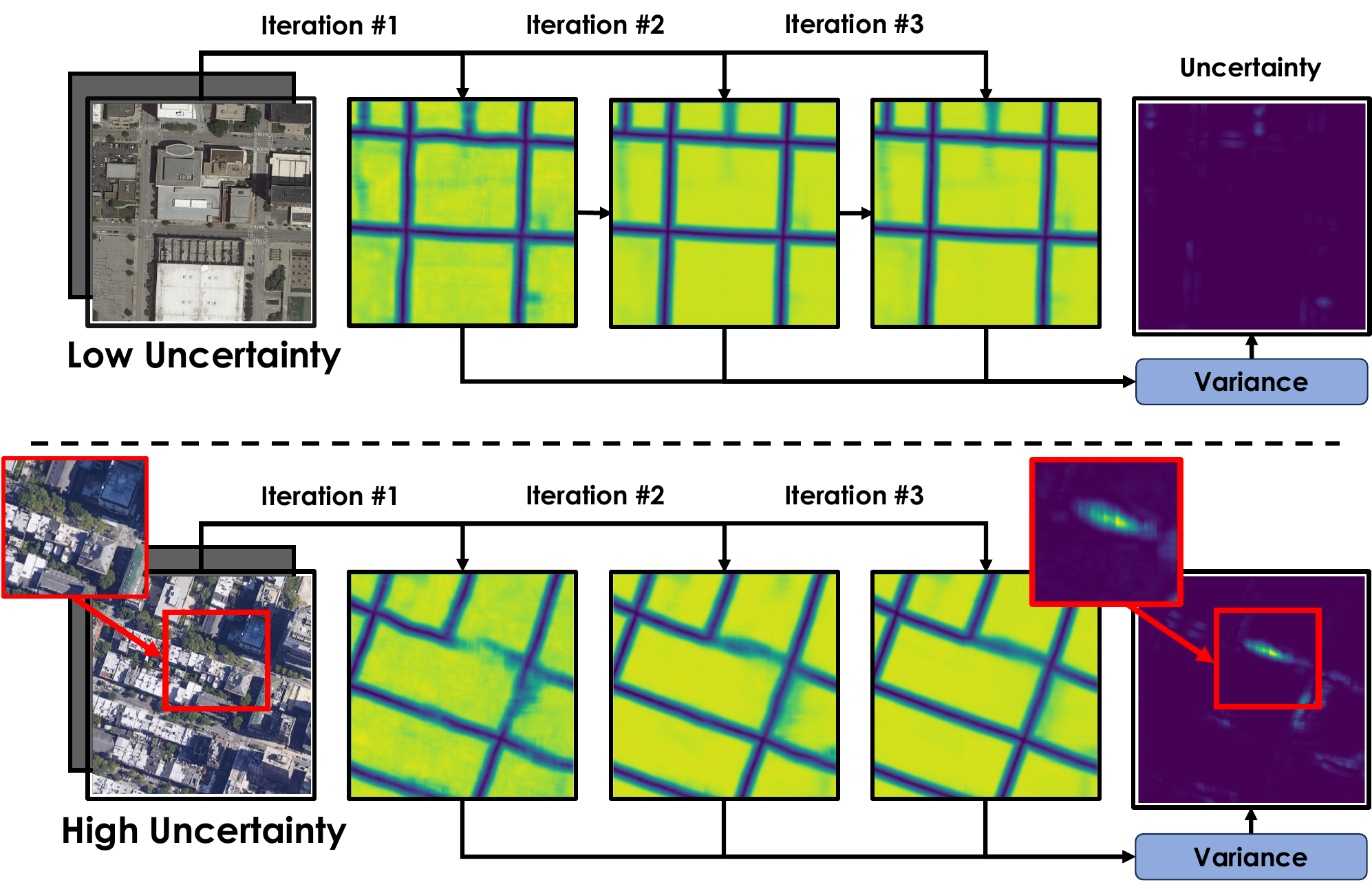}
    \vspace{-5mm}
    \centering
    \vspace{-3mm}
    \caption{\small 
    \textbf{Uncertainty in recursive models.} Such models use their initial predictions as inputs to produce subsequent predictions. We display the output of three consecutive iterations of a network trained to compute distance maps to road pixels.  \textbf{(Top:)} All roads are clearly visible. The three maps are similar and the per pixel variance is low.   \textbf{(Bottom:)} The road in the red square is tree-covered. It is eventually detected properly but the variance is high.}
    \label{fig:teaser}
\end{figure}

It has long been known that using deep networks to recursively refine predictions is often beneficial. This has been demonstrated for  semantic segmentation~\cite{Zhou18d, Wang19c}, pose estimation~\cite{Newell16}, depth estimation~\cite{Zhang18}, multi-task learning~\cite{Xu18c}, delineation~\cite{Mosinska18}, natural language processing~\cite{Vaswani17, Devlin18}, among others. Given a network $f$ parameterized by weights $\Theta$ and takes as input a vector $\bx$, which can represent an image or a text, and produces an output $\by$, the output of the recursion's $i^{\rm th}$ iteration can be written as $\by_{i} = f_{\Theta}(\bx,\by_{i-1})$, where $\by_{i-1}$ is the output of the previous iteration. This recursion often yields improved performance over non-iterative methods using the same network architectures~\cite{Shen17, Zhang18multi, Wang19c, Oner22a, Oner22b}, while requiring less labeled data for training purposes~\cite{Mosinska18}.  In essence, giving a previous output as input to the network sets up a virtuous circle in which the model receives relevant spatial attention signals that serve as priors and help generate improved predictions. For example, in the road delineation example of Fig.~\ref{fig:teaser}, the presence of road fragments with gaps in them cues the network to the possible existence of connecting segments.  These ideas been explored before the advent of Deep Learning, for example using Tensor Voting~\cite{Medioni00}, but incorporating them into deep networks has given them a new lease on life.

The key insight of this paper is that how fast this refinement occurs is closely connected to the accuracy of the prediction.  A hard sample typically requires more refinement iterations than an easy one.  Thus, convergence speed can be used as a proxy for certainty.  This yields an approach to uncertainty estimation that is on par with \textit{Deep Ensembles}~\cite{Lakshminarayanan17}, while delivering increased accuracy at a much lower computational cost and without requiring any modifications to the original iterative model. This makes our approach practical and easy to deploy across many different applications.  This is significant because Ensembles is often considered to be one of the very best uncertainty estimation methods whose only severe drawback is its high computational cost. 

More specifically, we derive uncertainty measures by analyzing the variance in outputs from consecutive iterations of an iterative model, where higher variance indicates greater uncertainty. We demonstrate that this is fast, accurate, and easy to deploy in two very different application domains, road detection in aerial images and the estimation of aerodynamic properties of 2D and 3D shapes. These two applications feature unique sets of challenges, which underscore the versatility and broad applicability of our approach. 


Our contributions are as follows:
\begin{itemize}

    \item We introduce an effective method for estimating uncertainty in iterative models. It relies on consistency across consecutive predictions and does not require modifying the network architectures. 
    
    \item We provide extensive experiments and analyses to demonstrate the correctness and effectiveness of the proposed method. In particular, we show that going from a toy example to our two complex real-world scenarios, there remains a consistent correlation between convergence speed and prediction accuracy.

    \item Our method, once embedded in a Bayesian Optimization framework, delivers state-of-the-art accuracy and predictive uncertainty quantification in both road detection from aerial images and 2D/3D shape optimization.
    
    
    
\end{itemize}
The code will be made publicly available.

\section{Related Work}

\subsection{Uncertainty Estimation}
\label{sec:related_unc}

Uncertainty Estimation (UE) aims at accurately evaluating the reliability of a model's predictions. Deep Ensembles~\citep{Lakshminarayanan17}, MC-Dropout~\citep{Gal16a}, and Bayesian Networks~\citep{MacKay95} have emerged as the most influential approaches. 

\ens{} involve training multiple networks, starting from different initial conditions. They are noted for the potential diversity of their predictions, attributable to randomness from weight initialization, data augmentation, and stochastic gradient updates. This diversity is central to their effectiveness~\cite{Perrone95,Fort19}. In many situations, they deliver more reliable uncertainty estimates than other methods~\cite{Ovadia19,Gustafsson20,Ashukha20,Postels22}. They therefore remain the leading technique, despite the high computational cost of training several networks instead of a single one and of performing several forward passes at inference time. One of the active research directions is reducing the training and inference time of ensembling methods, as well as their memory requirements. For example, \citet{antoran2020depth} try to emulate ensemble predictions by producing several outputs based on features at different levels of the model, and \citet{daxberger2021bayesian} perform Bayesian inference only on a subset of the model's weights chosen through a pruning-like procedure.

Among the other techniques, \mcdp{} involves randomly zeroing out network weights and assessing the effect and is popular due to its lower computational cost. Unfortunately, its estimates remain less reliable than those of \ens{}~\cite{Ashukha20}, even though there has been recent attempts at improving it~\citep{Wen2020, Durasov21a}. Similarly, {\it Bayesian Networks} rarely outperform \ens{}~\citep{Blundell15,Graves11,Hernandez15a,Kingma15b}.



All the above methods are sampling-based and require several forwards passes at inference time. Thus, when a fast response time is required, as in robotics control,  sampling-free approaches with single-pass inference become of interest. However, deploying them often requires significant modifications to the network's architecture~\cite{Postels19}, substantial changes to the training procedures~\cite{Malinin18}, limiting their application to very specific tasks~\cite{VanAmersfoort20, Malinin18, Mukhoti21a}, or reducing the quality of the uncertainty estimate~\cite{Postels22, Ashukha20}. As a result, they have not gained as much traction as MC-Dropout and Ensembles.

\subsection{Iterative Refinement Methods}
\label{sec:related_iter}

Iterative refinement techniques have been for many different purposes~\cite{Mnih10, Pinheiro14,Tu09,Shen17}. This incorporates surrounding context into the prediction~\cite{Seyedhosseini13}, proving particularly useful for tasks such delineation ~\cite{Sironi16a}, human pose estimation~\cite{Newell16}, semantic segmentation \cite{Zhang18multi,Wang19c}, depth estimation \cite{Durasov18,Xu18c}, and multi-task learning~\cite{durasov2022partal}. \hl{In the first use case of this work}, we build upon the recursive networks used in~\cite{Mosinska18} to delineate roads by computing distance maps to the road pixels, as shown in  Fig.~\ref{fig:teaser}. The network is a UNet~\cite{Ronneberger15} that takes as input the image and the distance map computed at the previous iteration, starting from a blank one. However, whereas the typical focus of delineation papers is to increase performance in terms of a number of delineation metrics, ours is to provide an uncertainty estimate on the detections without sacrificing performance. \hl{In the second use case, we use an iterative model with Graph Neural Networks~\cite{Monti17, Baque18} for 2D and 3D shape optimization. We show that uncertainty measures can be used to effectively select out-of-distribution data to enhance the training dataset.  }

\section{Method}
\label{sec:method}


Let us consider a {\it recursive} network $f_{\Theta}$, where $\Theta$ are the network weights. $f_{\Theta}$ takes as input a vector $\bx$---an image or a 3D shape in the examples presented in the results section---and its own output $\by$---a segmentation image or a pressure field in our examples. At the $i^{\rm th}$ iteration, we have
\begin{equation}
	\by_{i} = f_{\Theta}(\bx, \by_{i-1}) \; , \label{eq:recur}
\end{equation}
where the initial value $\by_0$ can be taken to be a vector of zeros. This computation is repeated $N$ times and $\by_N$ is taken to be the final output. In supervised approaches, the network is trained so that $\by_N$ matches the ground truth, with~\cite{Newell16, Carreira16} or without~\cite{Chen18g, Gupta20} supervision on the intermediate outputs $\{\by_1, \ldots \by_{N-1}\}$. Our key insight is that, at inference time, rather than treating $\by_N$ as the sole output, as is usually done, we should consider the whole sequence $\bY = \{ \by_1, \ldots \by_{N}\}$ because it provides valuable information about prediction certainty.

In the remainder of this section, we first discuss the behavior of this series of estimates and then propose a simple algorithm that exploits it to estimate uncertainty. 

\subsection{Motivation}
\label{sec:motivation}

Each iteration of Eq.\ref{eq:recur} takes the current prediction $\by_{i-1}$ and refines it into $\by_i$. This resembles what a denoising auto-encoder does when mapping a noisy input signal to its true value. Hence, the theoretical understanding of reconstruction errors in auto-encoders is relevant to our problem. 

\hl{In fact,} the reconstruction error for a sample fed into an autoencoder can indicate whether the sample lies within the training distribution of the model or not~\citep{Japkowicz95}. More formally, given a denoising or contractive autoencoder, ${\cal R}$, and a sample $\bx$, the reconstruction error $| {\cal R}(\bx)-\bx|$ is closely related to the log-probability of the data distribution $p_{data}(\bx)$~\cite{Bengio13generalized, Alain14}. This understanding was later expanded to include a broader spectrum of autoencoders \citep{Kamyshanska13}, and then to standard autoencoders trained under stochastic optimization~\citep{Solinas20}. This theoretical work has been effectively applied to many practical tasks requiring out-of-distribution detection \citep{Zhou22rethinking, Sabokrou16video}.

\begin{figure}[htb]
    \centering
    \begin{tabular}{c}
        \includegraphics[width=0.47\textwidth]{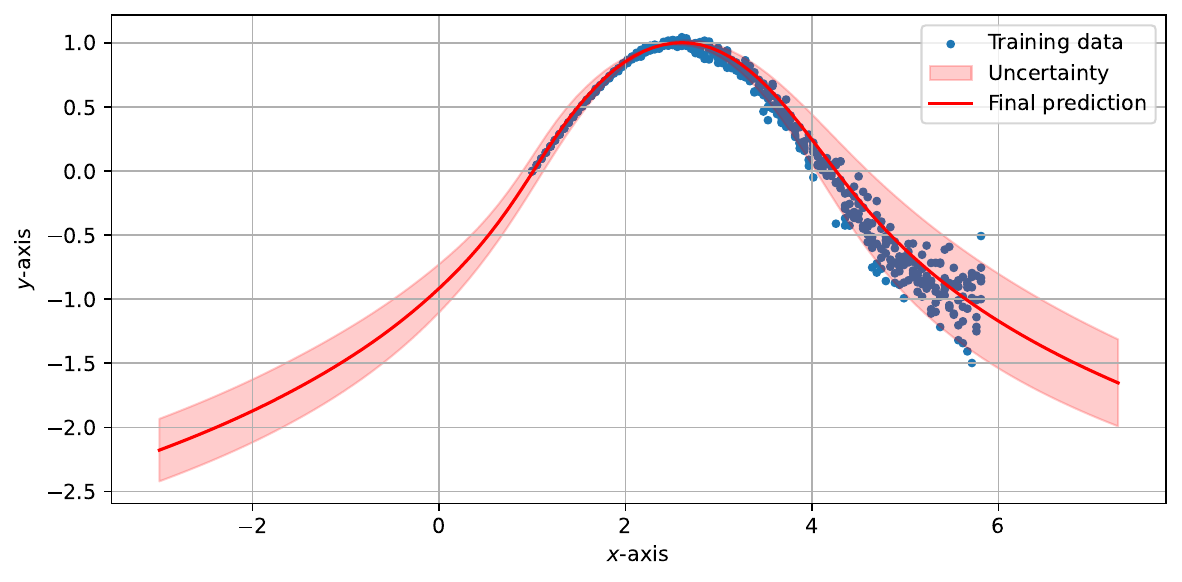} \\[-2mm]
        \includegraphics[width=0.47\textwidth]{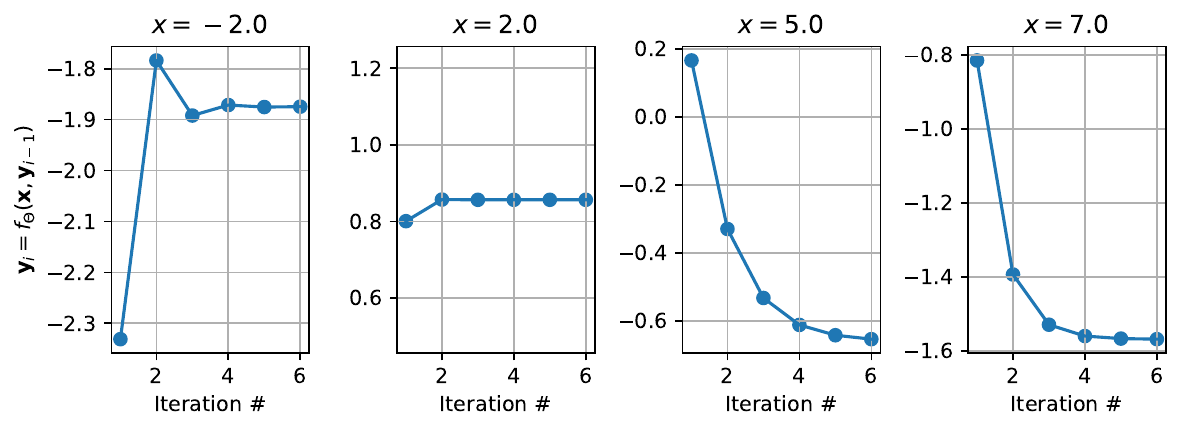}
    \end{tabular}  
     \vspace{-6mm}   
    \caption{\small \textbf{Uncertainty vs Convergence.} In this example,  we generated training data from a sinusoidal function for $x \in [1, 6]$ and added Gaussian noise with a variance that increases from left to right. We take $f_{\Theta}$ to be a simple MLP with three hidden layers that takes two inputs, $x$ and the output of the previous iteration. We train it to predict the noisy data points at each step of the iterative process by minimizing the loss of Eq.~\ref{eq:loss}. Once trained, we use $f_{\Theta}$ to produce predictions $\bY(x) = \{ \by_1(x), \ldots \by_{N}(x)\}$ for $x \in [-3, 7]$ \textbf{(Top):} The red line denotes the final prediction $\mathbf{y}_{N}(x)$, and the standard deviation of $\mathbf{Y}(x)$ is shown in pink. It increases away from the data and when the data is noisy, as it should.  \textbf{(Bottom):} The plots depict the values in the sequence  $\bY(x)$ for four different values. For $x=2$, both aleatoric and systemic uncertainties are low and convergence quickly. For $x=5$, the aleatoric uncertainty data is high because the data is noisy and the convergence is slow. For $x=-2.0$ and $x=7.0$ the systemic uncertainty is high because the points are out of distribution and the convergence is slow or erratic. }
    \label{fig:theory}
\end{figure}

Thus, as in~\citet{Bengio13generalized, Alain14}, we can rewrite the update equation of Eq.~\ref{eq:recur} as
\begin{equation}
    \mathbf{y}_{i+1} -  \mathbf{y}_{i} = f_{\Theta}(\mathbf{x}, \mathbf{y}_{i}) -  \mathbf{y}_{i}  \propto \frac{\partial \log p(\mathbf{y}_{i} | \bx)}{\partial \mathbf{y}_{i}} \; , \label{eq:autoenc}
\end{equation}
where $p(\mathbf{y}_{i} | \mathbf{x})$ is the probability of the model yielding the prediction $\by_{i}$ given the input $\bx$. In other words, the recursion can be understood as a gradient ascent on $\log p(\mathbf{y}_{i} | \mathbf{x})$, which explains why the prediction progressively improves as illustrated by Fig.~\ref{fig:theory} in a simple regression case. We can distinguish three different scenarios
\begin{itemize}

\item {\bf In distribution, without aleatoric noise.} $\bx$ is in-distribution, there is little noise in the training data, which makes the {\it aleatoric uncertainty} low and the probability $p$ peaked. So, if $\by_{i}$ is already close to being correct, the derivative of the log probability will be small and it will not move much. In contrast, it is not correct but still within the main mode of the probability distribution, the derivative  will be large and the convergence rapid. This is the case $x=2.0$.

\item {\bf In distribution, with aleatoric noise.} $\bx$ is in-distribution, the training data is noisy, making the  he {\it aleatoric uncertainty} higher and the probability $p$ less peaked. If $\by_{i}$ is not initially correct  but still within the main mode of the probability distribution, the derivative will be smaller than in the no-noise scenario, and thus the convergence slower, as in the case of $x=5.0$.

\item {\bf Out of distribution.} $\bx$ is out-of-distribution, which can be understood as {\it epistemic uncertainty}  and $p$ cannot be expected to have a well-defined peaked shape but often tends to be flatter. The behavior is then somewhat random as in the case of $x=-2.0$ and $x=7.0$.

\end{itemize}
In short, both aleatoric and epistemic uncertainty are likely to impact convergence speed negatively. 


\subsection{Estimating Uncertainty}
\label{sec:estimating}

In Section~\ref{sec:motivation}, we have argued that when the input $\bx$ to network $f_{\Theta}$ is in-domain with respect to the sample distribution that was used to train the network and the aleatoric uncertainty is low, we can expect the convergence of sequence $\bY$ tends to be quick. In contrast, when $\bx$ is out-of-domain or the aleatoric uncertainty is high, we can expect it to be far more erratic. To exploit this insight, for a given $\bx$ t we take the variance of sequence $\bY$ as a proxy for uncertainty.  We take it to be
\begin{equation}
    U^{i} = Var(\{\by_1^i, \by_2^i, \cdots, \by_N^i\}) \; , \label{eq:variance}
\end{equation}
where $i$ corresponds to the $i^{\rm th}$ pixel/node in the original input $\bx$ \hl{and $N$ is the number of iterations passing through the iterative network}. 
\hl{Here the variance represents the convergence speed, where higher values imply slower convergence and vice versa.} This approach lets us evaluate uncertainties at the pixel or node level. To obtain a scalar uncertainty estimate for the whole output $\by$, we average the values across all pixels/nodes.

In the result section, we show that this variance estimate strongly correlates with the actual accuracy of the prediction on experimental data. Note that for the arguments made for two in-distribution cases discussed at the end Section~\ref{sec:motivation} to apply, the prediction $\by_i$ has to fall within the mode of the probability distribution. To maximize the chances of this happening,  at training time, we supervise the network so that all $\by_i$ in $\bY$ are as close possible to the ground-truth by minimizing
\begin{equation}
    \mathcal{L}_{total} = \sum_{i=1}^{N}  \mathcal{D}(\by_{i}, \by^{\rm gt}) \label{eq:loss},
\end{equation}
where $D$ is a measure of distance and gt stands for ground-truth. 

\section{Experiments}
\label{sec:experiments}

Our approach applies to both classification and regression. To demonstrate the first, we use it for road delineation purposes, that is, classifying pixels in aerial images as belonging to roads or not. To demonstrate the second, we use it to assess the reliability of performance numbers----drag for cars and lift-to-drag for airfoils---predicted by networks given 2D and 3D shapes as input. We then use these reliability estimates to implement a Bayesian optimization scheme that enables us to refine the shapes for improved performance. For both classification and regression, we outperform Deep Ensembles and MC-Dropout, along with Kriging in the regression case. 

\subsection{Delineation}
\label{sec:classif}

Tasks such as road detection or modeling thin biological structures from images fall within the heading of visual delineation. After more than 50 years of research, it remains an open topic even though modern networks have boosted the state-of-the-art. Their final output often is a binary map indicating where in the image pixels belonging to structures of interest are. Generating such a map can be viewed as classifying the pixels as belonging to the target structures or not. 




\parag{Datasets.}

We experimented on two publicly available datasets.
\begin{itemize}

\item \textit{RoadTracer}. It comprises high-resolution satellite images covering urban areas of forty cities in six different countries~\cite{Bastani18}. Fifteen cities are set aside for validation purposes. The ground truth was generated using OpenStreetMap.

\item \textit{Massachusetts}. The Massachusetts dataset features both urban and rural neighborhoods, with many different kinds of roads ranging from small paths to highways. We used the same splits as in~\cite{Hu19b}.

\end{itemize}
Together, these datasets exhibit a very large variety of urban scapes, which makes them a comprehensive benchmark for aerial road network reconstruction.

\parag{Baselines.}

Architectures such as U-Net~\cite{Ronneberger15} or SegNet~\cite{Badrinarayanan15} are commonly employed for delineation purposes, given their effectiveness in image segmentation challenges.  For a fair comparison, all the method we tested rely on the standard U-Net~\cite{Ronneberger15} architecture, with five blocks, each with three sequences of convolution-ReLU-batch normalization. Max-pooling in $2\times2$ windows followed each of the blocks. The initial feature size was set to $32$ and grew to $1024$ in the smallest feature map in the network. The network is trained to output a distance map that can then be thresholded to produce the binary one. We augmented the training data with vertical and horizontal flips, along with  random rotations. Thus, the four methods we compare are
\begin{itemize}

\setlength\itemsep{1.4mm}
 
\item \unet{}. Standard U-Net~\cite{Ronneberger15}. 

\item \mcdp{}. Adding drop-out layers~\cite{Gal16a} into the  standard U-Net  to estimate the mean and variance of the predictions.

\item \ens{}. Using five standard \textit{U-Net}s to estimate the mean and variance of the predictions~\cite{Lakshminarayanan17}.
 
\item \ours{}. Using a recursive version of the standard U-Net~\cite{Wang19c}, A dual-gated recurrent unit has been added in the bridge part of the network.  During training, we performed three iterations.  After each one, the output of the network is used as an additional input channel for the next one.

\end{itemize}
%
%


\begin{table}[ht!]
    \centering
    \begin{small}
    \begin{tabular}{c|ccc|cc|c} 
    &  \textit{Corr} & \textit{Comp} & \textit{Qual} & \textit{F1} & \textit{APLS} & \\
    \hline
    \unet & 85.2 & 59.5 & 54.3 & 21.1 & 65.04 & \multirow{4}{*}{\rotatebox[origin=c]{270}{RT}} \\ 
    \mcdp & \cellcolor{secondbest}\graybold{87.1} & 58.2 & 54.1 & 20.4 & 58.78 \\
    \ens & \cellcolor{best}\textbf{87.4} & \cellcolor{secondbest}\graybold{66.7} & \cellcolor{secondbest}\graybold{60.8} & \cellcolor{secondbest}\graybold{22.1} & \cellcolor{secondbest}\graybold{68.81} \\
    \ours & 85.2 & \cellcolor{best}\textbf{77.8} & \cellcolor{best}\textbf{68.6} & \cellcolor{best}\textbf{24.5} & \cellcolor{best}\textbf{77.21} \\
    \hline
    \unet  & 81.5 &\cellcolor{secondbest}\graybold{91.4} & 77.8 & 13.8 & 65.42 & \multirow{4}{*}{\rotatebox[origin=c]{270}{MS}} \\ 
    \mcdp & 81.6 & \cellcolor{best}\textbf{92.3} & 78.2 & 13.6 & 59.65 &\\
    \ens & \cellcolor{secondbest}\graybold{83.6} & 90.4 & \cellcolor{secondbest}\graybold{78.7} & \cellcolor{secondbest}\graybold{14.1} & \cellcolor{secondbest}\graybold{67.53} &\\
    \ours& \cellcolor{best}\textbf{92.3} & 86.7 & \cellcolor{best}\textbf{81.1} & \cellcolor{best}\textbf{15.4} & \cellcolor{best}\textbf{78.04} &\\
    \end{tabular}
     \end{small}
    \vspace{-2mm}
    \caption{\small {\bf Delineation accuracy on \textbf{RoadTracer} (top), and \textbf{Massachusetts} (bottom).} The best result in each category is  in {\bf bold} and the second best is in \graybold{bold}. Most correspond to \textit{Ours} and {\it DeepE}.}
    \label{tab:acc_results}  
\end{table}

\parag{Metrics.}


For road delineation, the true measure of success is preservation of the topology of the road networks rather than the very precise location of the centerline. This is ussually expressed in terms of hte following metrics: 
\begin{itemize}

    \setlength\itemsep{2mm}
     \item \textit{APLS} ($\uparrow$).  Average Path Length Similarity, defined as an aggregation of relative length difference of shortest paths between pairs of corresponding points in the reconstructed and predicted maps~\cite{van2019spacenet}.

     \item \textit{CCQ} ($\uparrow$). Metric that measures spatial co-occurrence of annotated and predicted road pixels. The \textit{Correctness}, \textit{Completeness} and \textit{Quality} are equivalent to precision, recall and intersection-over-union, where the definition of a true positive has been relaxed from spatial coincidence of prediction and annotation to co-occurrence within a distance of 5 pixels~\cite{Wiedemann98}.
    
     \item \textit{F1 Score} ($\uparrow$). A balance between precision and recall, the F1 score is twice the product of precision and recall divided by their sum. It's widely used in binary segmentation to equally weigh false positives and false negatives~\cite{Fawcett06}.
     
\end{itemize}
To similarly evaluate the quality of the uncertainty estimates, as in~\citep{Postels22}, we compute 
\begin{itemize}

    \setlength\itemsep{2mm}
     \item \textit{Relative Area Under the Lift Curve} (rAULC). It is derived from the \textit{Area Under the Lift Curve} concept~\citep{Vuk06} and assesses the calibration quality of uncertainty measures across various methods. 

     \item \textit{Pearson Correlation Coefficient} (Corr). It measures the correlation between the estimated uncertainty and the actual error. 
     
\end{itemize}

\begin{figure}[htb]
    \centering
    \begin{tabular}{ccc}
        \hspace{-2mm}\includegraphics[width=0.15\textwidth]{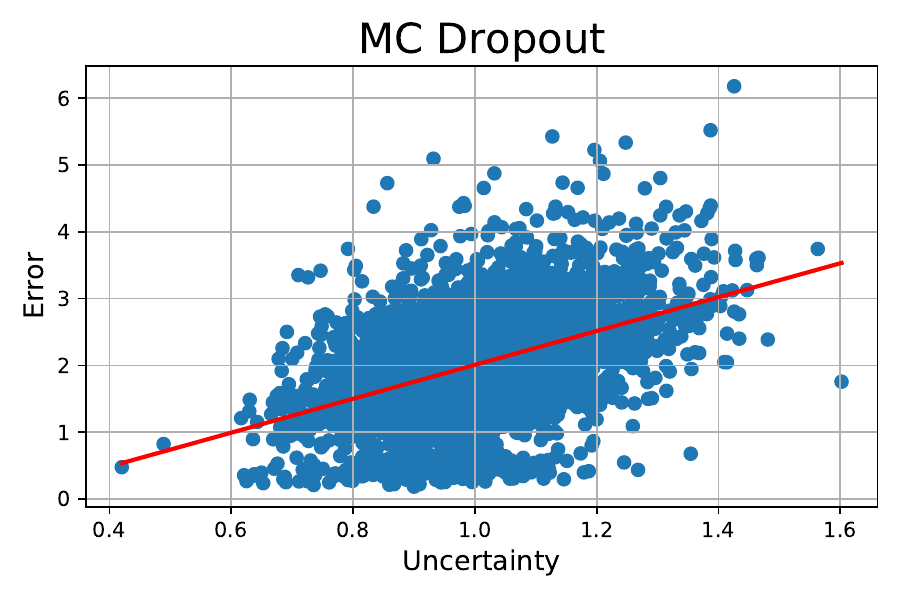}&
        \hspace{-2mm}\includegraphics[width=0.15\textwidth]{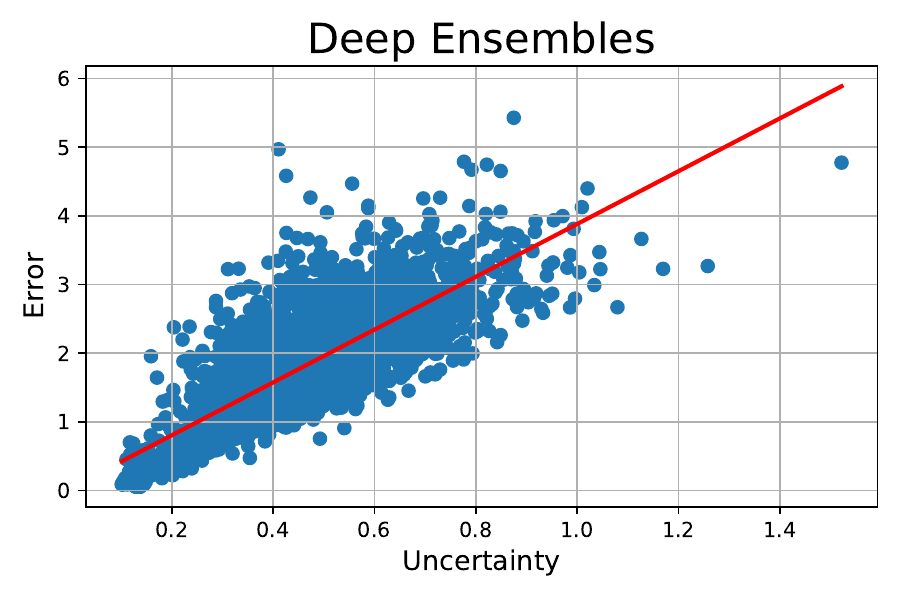}&
        \hspace{-2mm}\includegraphics[width=0.15\textwidth]{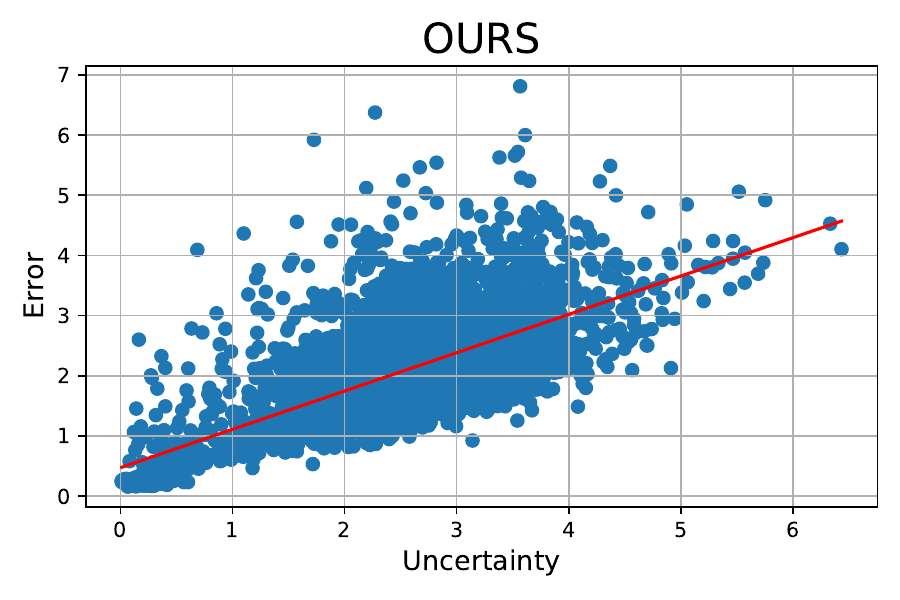}\\[1mm]
        \hspace{-2mm}\includegraphics[width=0.15\textwidth]{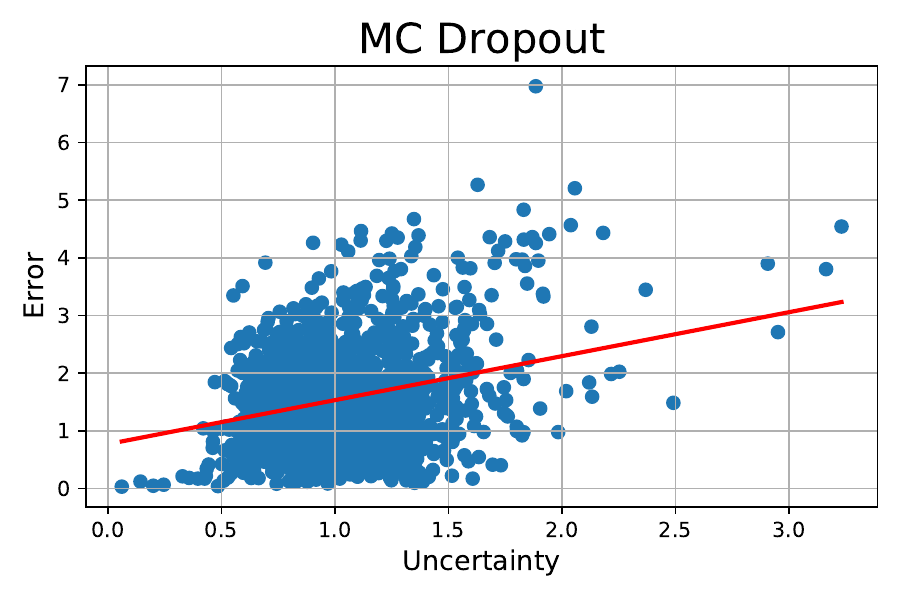}&
        \hspace{-2mm}\includegraphics[width=0.15\textwidth]{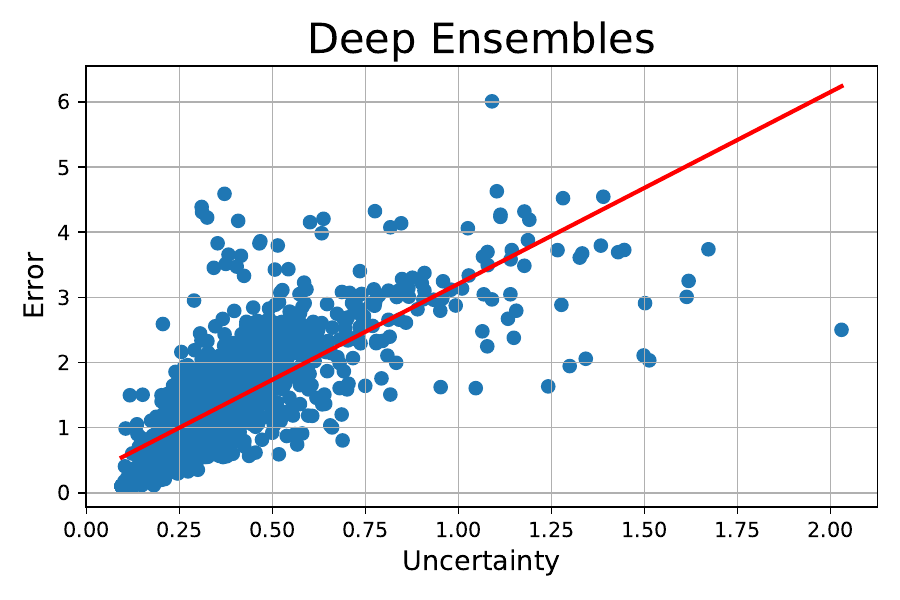}&
        \hspace{-2mm}\includegraphics[width=0.15\textwidth]{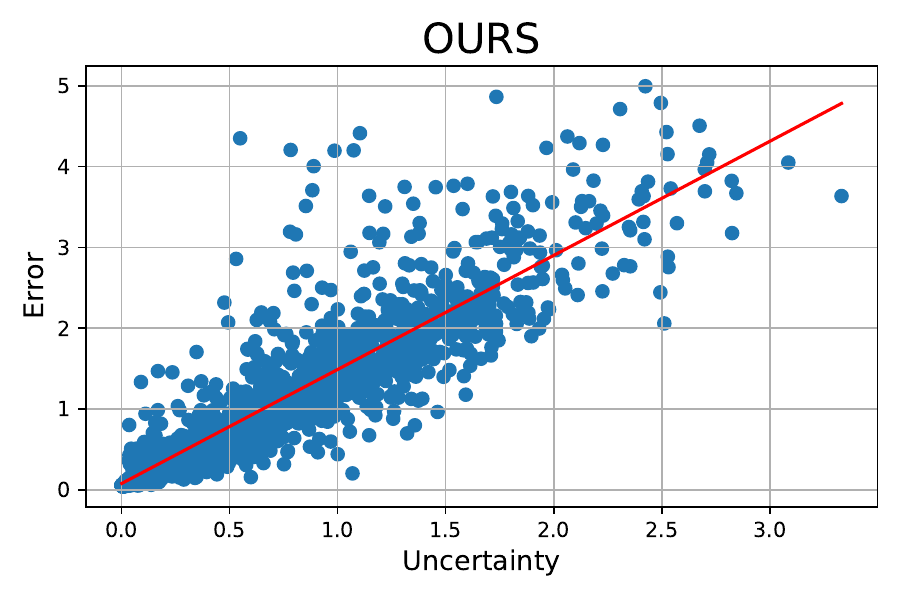}\\[1mm]
    \end{tabular} 
    \vspace{-6mm}    
    \caption{\small 
        \textbf{Error vs Uncertainty.} These plots illustrate the error-uncertainty relationship for three methods on the \textit{RoadTracer} \textbf{(Top)} and \textit{Massachusetts} \textbf{(Bottom)} datasets. Our method surpasses the others on the \textit{Massachusetts} dataset and performs comparably with Ensembles on \textit{RoadTracer}. Correlation numbers are in Tab.~\ref{tab:unc_results_delineation}. The red line indicates the optimal linear fit.}
    \label{fig:delineation_corr}
\end{figure}

\parag{Evaluation.}
We report accuracy and uncertainty results in Tabs.~\ref{tab:acc_results} and~\ref{tab:unc_results_delineation}. For all uncertainty evaluations, we calculate predictions and uncertainty for each pixel in the image. We then divide these into $512\times512$ crops, averaging the uncertainties and errors across each crop to ensure a more stable evaluation of the metrics. In terms of uncertainty estimation, \ens{} and \ours{} are comparable and outperform the others.  In terms of accuracy,  \ours{} does best. To highlight this, in Fig.~\ref{fig:delineation_corr}, we show scatter plots of estimated uncertainty vs actual accuracy. Note that our results exhibit a more linear behavior, which is what the Pearson Correlation Coefficient measures.



\begin{table}[ht!]
    \centering
    \begin{tabular}{c|cc|cc|c} 
    & \textit{rAULC} & \textit{Corr} & \textit{Train} & \textit{Inf}  \\
    \hline
    \mcdp & 30.18 & 59.72 & \cellcolor{best}\textbf{1x} & \cellcolor{secondbest}\graybold{5x} &\multirow{3}{*}{\rotatebox[origin=c]{270}{RT}} \\ 
    \ens & \cellcolor{best}\textbf{72.19} & \cellcolor{best}\textbf{79.42} & 5x & \cellcolor{secondbest}\graybold{5x} &  \\
    \ours &  \cellcolor{secondbest}\graybold{69.23} & \cellcolor{secondbest}\graybold{74.73} & \cellcolor{secondbest}\graybold{2.8x} & \cellcolor{best}\textbf{2.7x} & \\
    \hline
    \mcdp & 19.56 & 32.50 & \cellcolor{best}\textbf{1x} & \cellcolor{secondbest}\graybold{5x} &  \multirow{3}{*}{\rotatebox[origin=c]{270}{MS}} \\ 
    \ens & \cellcolor{secondbest}\graybold{78.65} &  \cellcolor{secondbest}\graybold{76.39} & 5x & \cellcolor{secondbest}\graybold{5x} & \\
    \ours & \cellcolor{best}\textbf{79.27} & \cellcolor{best}\textbf{87.46} & \cellcolor{secondbest}\graybold{2.8x} & \cellcolor{best}\textbf{2.7x} & \\
    \end{tabular}
    \caption{\small {\bf Delineation uncertainty quality on \textbf{RoadTracer} (top), and \textbf{Massachusetts} (bottom).} The best result in each category is  in {\bf bold} and the second best is in \graybold{bold}. Most correspond to \textit{Ours} and {\it DeepE}. The \textit{Train} and \textit{Inf} metrics represent the total training time and inference time for the model, respectively, relative to a single model.}
    \label{tab:unc_results_delineation}  
\end{table}

\begin{table}[ht!]
    \centering
    \begin{tabular}{c|c|c|c} 
    & \textit{ROC-AUC} & \textit{PR-AUC} & \\
    \hline
    \mcdp & 61.25  & 62.64 &\multirow{3}{*}{\rotatebox[origin=c]{270}{RT}} \\ 
    \ens & \cellcolor{secondbest}\graybold{67.03} & \cellcolor{secondbest}\graybold{67.85} & \\
    \ours & \cellcolor{best}\textbf{67.09} & \cellcolor{best}\textbf{72.11} &\\
    \end{tabular}
    \caption{\small {\bf \textbf{RoadTracer vs Massachusetts} out-of-distribution detection results.} The best result in each category is  in {\bf bold} and the second best is in \graybold{bold}. Most correspond to \textit{Ours} and {\it DeepE}}
    \label{tab:ood_results}  
\end{table}

To further evaluate our uncertainty estimates, we use the same insight as in~\citep{Malinin18, Durasov22}: A network trained on samples drawn from a given distribution should be more confident on samples drawn from the same distribution than on samples drawn from a different one. 
\nd{
In this context, we conducted an out-of-distribution detection task. We utilized the model trained on the \textit{RoadTracer} dataset, treating its test set as in-distribution data. For out-of-distribution data, we selected the test set of the \textit{Massachusetts} dataset. These two datasets exhibit markedly different landscapes. \textit{RoadTracer} primarily features images of urban centers, whereas \textit{Massachusetts} encompasses aerial images of rural areas.
}

We rely on the uncertainty measure generated by our model to decide whether a sample is in-domain or out-of-domain.  We then apply standard detection metrics, ROC and PR AUCs~\citep{Malinin18}, to quantify the performance of our model. As for calibration metrics, we perform this evaluation using $512\times512$ crops: \nd{we average per-pixel uncertainties across the entire crop and classify it as in- or out-of-distribution based on this averaged uncertainty value.} We report our results in Tab.~\ref{tab:ood_results}.

\subsection{Aerodynamics Prediction and Optimization}
\label{sec:regression}


\begin{figure*}[htb]
    \includegraphics[width=0.9\textwidth]{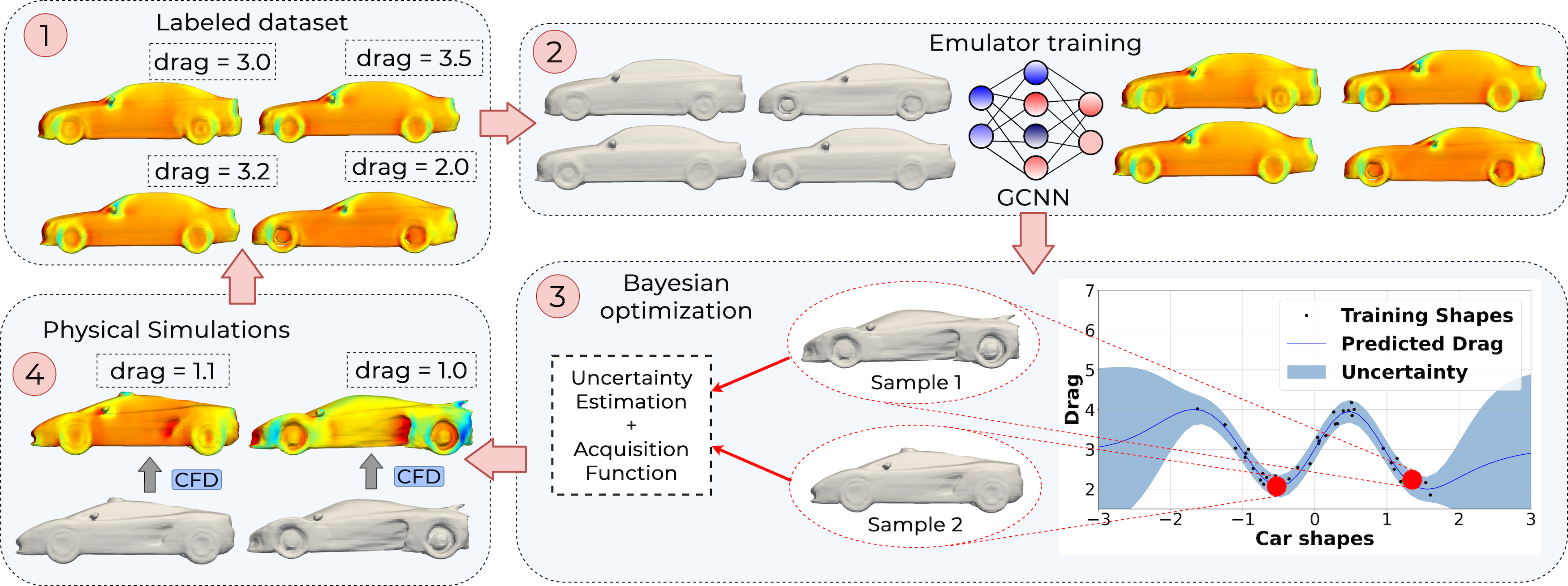}
    \vspace{-2mm}
    \centering
    \caption{\small 
    \textbf{Bayesian optimization pipeline.}
    \textbf{(1)} Run physical simulations.
    \textbf{(2)} Train the GNN. 
    \textbf{(3)} Evaluate the acquisition function on samples without an associated physical simulation. 
    \textbf{(4)} Select promising samples according the acquisition function,  optimize their shape, add them to the training set, and go back to step 1. 
    }
    \vspace{-2mm}
    \label{fig:pipeline}
\end{figure*}

We now showcase the effectiveness of our approach at  estimating the uncertainly of surrogate models used to estimate and optimize aerodynamic performance.



\paragraph{Bayesian optimization.} 
To refine 2D and 3D shapes and increase their expected aerodynamic performance, we rely on Bayesian Optimization (BO)~\cite{Mockus12} as depicted by Fig.~\ref{fig:pipeline}. Our implementation comprises four steps: 
\begin{enumerate}[left=0pt] 
  \item Use the training shapes and simulation results to train a surrogate model $f_{\Theta}$.
  
  \item Take each shape from the unlabelled pool and make a prediction with $f_{\Theta}$.
  
  \item Given the uncertainty of the predictions, compute the \textit{acquisition function}~\cite{Auer02, Qin17} for the shapes in the unlabelled pool. This function balances between exploration and exploitation, and it is the key to the success of Bayesian optimization.
  
  \item Pick the best new shapes in terms of the acquisition function, optimize their shape with gradient optimization, add them to the training set, and iterate.
\end{enumerate}
These are standard BO steps, as described in Appendix~\ref{sec:bo}, except for step \#4. It involves exploring the shape space without running additional simulations. It takes advantage of the fact that GNNs allow for gradient-based shape optimization. The key to implementing  Bayesian shape optimization is an effective way to estimate not only the performance value associated with a shape but also the uncertainty on this estimate in step \#3, which is something ordinary GNNs~\citep{Monti17} do not provide, and which is being addressed by our approach.

\paragraph{Datasets.} Given a set of $N$ 3D shapes $\{\textbf{x}_i\}_{1 \leq i \leq N}$ represented by triangulated meshes, we run a physics-based simulator yielding a corresponding set  $\{\textbf{y}_i\}_{1 \leq i \leq N}$ of physical values, such as pressure at each vertex. Let $R$ be the function that takes as input the $\by$ values and returns an overall performance value  $r = R(\by)$, such as overall drag for a car or lift for a wing. $R$ is task-specific. For example, in the case of drag, it is computed by integrating pressure values over the 3D shape. The simulator also generates $\{r_{i}\}_{1 \leq i \leq N}$ in conjunction with the $\by_{i}$'s. Assuming that each mesh $\bx_{i}$ is parameterized by a lower-dimensional latent vector $\bz_{i}$ and that there is a differentiable mapping $\textbf{P}: \bz \rightarrow \bx$, this gives us the initial training set $T = \{(\bz_{i}, \bx_{i}, r_{i}, \by_{i})\}_{i}$ that we need to initialize our optimization scheme. Similarly, we expect a larger pool of unlabeled shapes, consisting of latent vectors and meshes denoted as $U = \{(\bz_{i}, \bx_{i})\}_{i}$, but no simulation data. We train the surrogate model using samples from $T$ (Step 1) and use it to perform predictions (Step 2), compute the acquisition function (Step 3), and select samples for simulations from the set $U$ (Step 4).
\begin{itemize}
    \item \textit{Airfoils}. We generated a dataset comprising 1500 two-dimensional airfoil shapes. This was achieved by randomly selecting NACA parameters, $\mathbf{z}_{i}$, and producing corresponding airfoil contours, $\mathbf{x}_{i}$. The pressure distribution, $\mathbf{y}_{i}$, over each airfoil surface was computed using the XFoil simulator. Additionally, the global lift-to-drag ratio, $\mathbf{r}_i$, a measure of aerodynamic efficiency, was calculated for each shape. The dataset was divided into 1000 training samples, 300 testing samples, and 200 high-performance shapes, treated as out-of-distribution samples for uncertainty analysis.
    
    \item \textit{Cars}. We use a cleaned-up and processed subset of the ShapeNet dataset~\citep{Chang15} that features $N = 1500$ car meshes suitable for CFD simulation. For each such mesh $\textbf{x}_{i}$, we run OpenFOAM~\citep{Jasak07} to estimate the pressure field $\textbf{y}_{i}$ and drag $\textbf{r}_{i}$ created by air traveling at 15 meters per second towards the car. We also use MeshSDF~\citep{Remelli20b} in conjunction with an auto-decoding approach~\citep{Park19c} to learn a function $P: \mathbb{R}^{256} \rightarrow  \mathbb{R}$ and a set of latent vectors $\{\bz_i\}$ such that $\forall i \; \bx_i = P(\bz_i)$. We use the same protocols as for \textit{Airfoils} for splitting. 
 
\end{itemize}

\paragraph{Baselines.} We compare our method against widely recognized and universally adopted baselines, which are considered the gold standard in the field:

\begin{itemize}

\item  \textit{KNN}: Given a set of simulated shapes, we use a standard K-Nearest Neighbors regressor to estimate the performance of additional shapes and add the best one to the training set. No uncertainty is computed. For this approach, we use $K=8$ and employ distance-based neighbor weighting, as this has been shown to be the optimal choice for this task~\citep{Baque18}.

\item  \textit{Kriging}: Using a Gaussian Processes (GPs) to estimate performance values and corresponding uncertainty~\citep{Laurenceau10} directly from parameters $\bz$. As discussed above, it can be directly used to perform Bayesian Optimization. For GPs, we use the squared exponential kernel,  which has been shown to be particularly effective for aerodynamic prediction~\citep{Toal11,Rosenbaum13}.
  
\item  \textit{GNN}: GNNs~\citep{Baque18,Hines22} are a valid alternative to GPs for the purpose of estimating performance numbers. Since they do not compute uncertainties, we simply add the ones that receive the best score from the GNN to the training set and optimize their shape as in~\citep{Baque18}. 
  
\item  \textit{Deep Ensembles}: We use sets of GNNs to predict mean and variances of performance values, which is known as an Ensemble-based technique. These are then exploited by the procedure introduced previously in this section. For all of the experiments, we use 5 GNNs in ensemble.
 
\item  \textit{MC-Dropout}: Instead of using Ensembles to estimate the performance numbers and their uncertainty, we use MC-Dropout in the  Bayesian optimization procedure.

\item \textit{Ours}: Using the iterative GNN to simultaneously estimate the performance values and their uncertainty for the  Bayesian optimization procedure.

\end{itemize}


\paragraph{Metrics.} We use the following metrics to evaluate the quality of our baselinse in terms of predictions accuracy and uncertainty estimation:
\begin{itemize}
    \item \textit{Mean absolute error} ($\downarrow$) (MAE). It is the average of the absolute differences between the predicted and actual values. It is a common metric for regression tasks.
    \item \textit{Opimized performance}. We use our baselines to perform Bayesian optimization as it was described above. We then report the best performance value, $\textbf{r}_{i}$ lift-to-drag value for airfoils and $\textbf{r}_{i}$ drag value for cars, obtained by each method and the dynamics of optimization process. 
\end{itemize}

\paragraph{Evaluation.}

\begin{table}[ht!]
    \centering
    \begin{tabular}{c|cccc|c} 
    &  \textit{Krig} & \textit{MC-DP} & \textit{DeepE} & \textit{Ours} & \\
    \hline
    ROC-AUC & $0.79$ & 0.84 & \cellcolor{best}\textbf{0.88} & \cellcolor{secondbest}\graybold{0.87} & \multirow{2}{*}{\rotatebox[origin=c]{270}{AIR}} \\ 
    PR-AUC & $0.78$ & $0.82$ & \cellcolor{secondbest}\graybold{0.86} & \cellcolor{best}\textbf{0.88} & \\

    \hline
    ROC-AUC & $0.62$ & $0.73$ & \cellcolor{best}\textbf{0.90} & \cellcolor{secondbest}\graybold{0.86} & \multirow{2}{*}{\rotatebox[origin=c]{270}{CAR}} \\ 
    PR-AUC & $0.52$ & $0.62$ & \cellcolor{secondbest}\graybold{0.78} &  \cellcolor{best}\textbf{0.79} & \\
    \end{tabular}
    \caption{\small {\bf Evaluation the uncertainty measure for 2D airfoils (\textbf{AIR}) and 3D cars (\textbf{CAR}). } The best result in each category is  in {\bf bold} and the second best is in \textcolor{gray}{\textbf{bold}}. They all correspond to \ours{} and \ens{}. The two approaches are comparable in terms of evaluating uncertainty but the reentrant GNNs deliver better accuracy, as shown in Fig.~\ref{fig:comp}.}
    \label{tab:aero_unc_results}  
\end{table}


%

\begin{figure*}[htb]
    \centering
    \begin{tabular}{cc|cc}
        \hspace{-4mm}\includegraphics[width=0.24\textwidth]{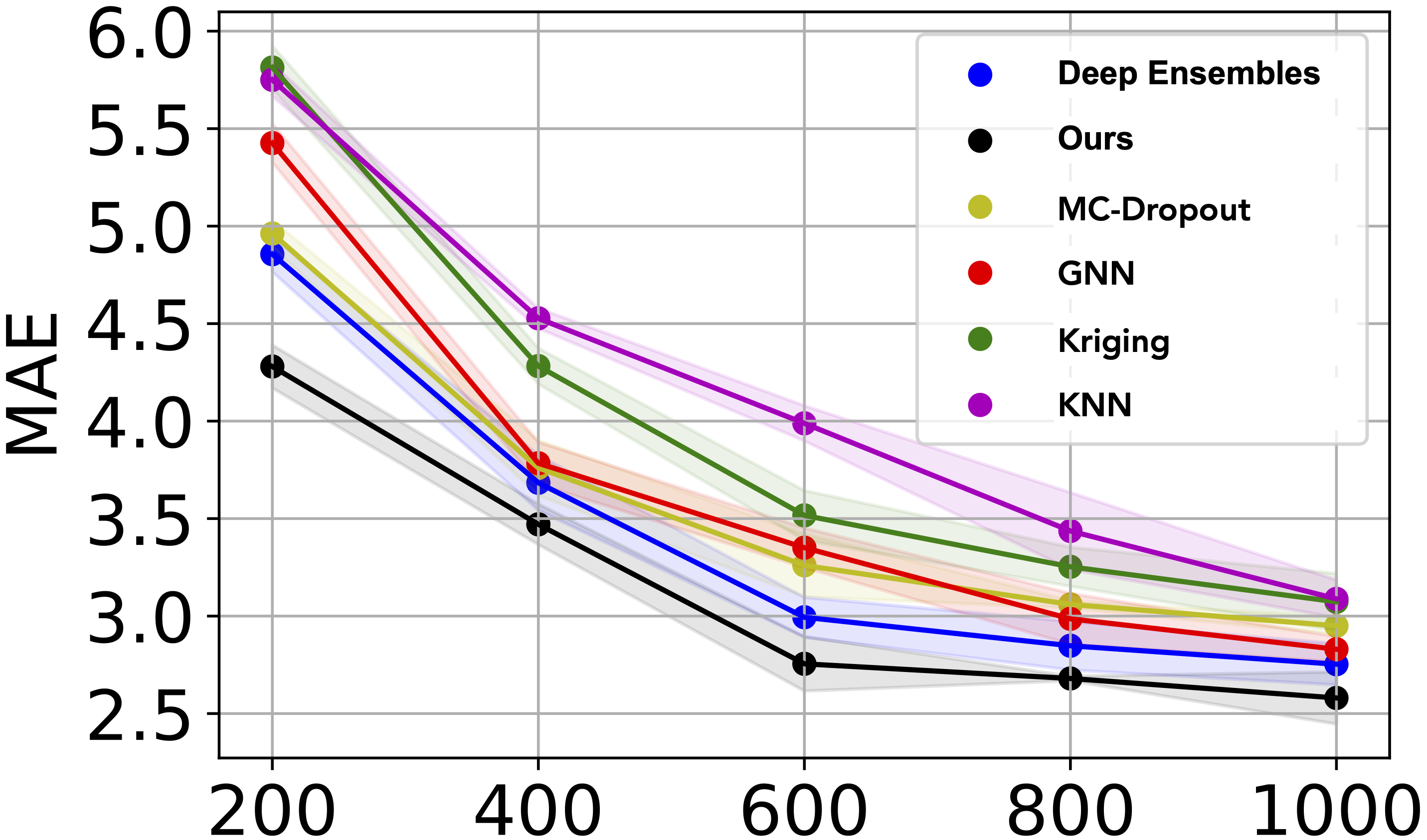} & \hspace{-4mm}\includegraphics[width=0.24\textwidth]{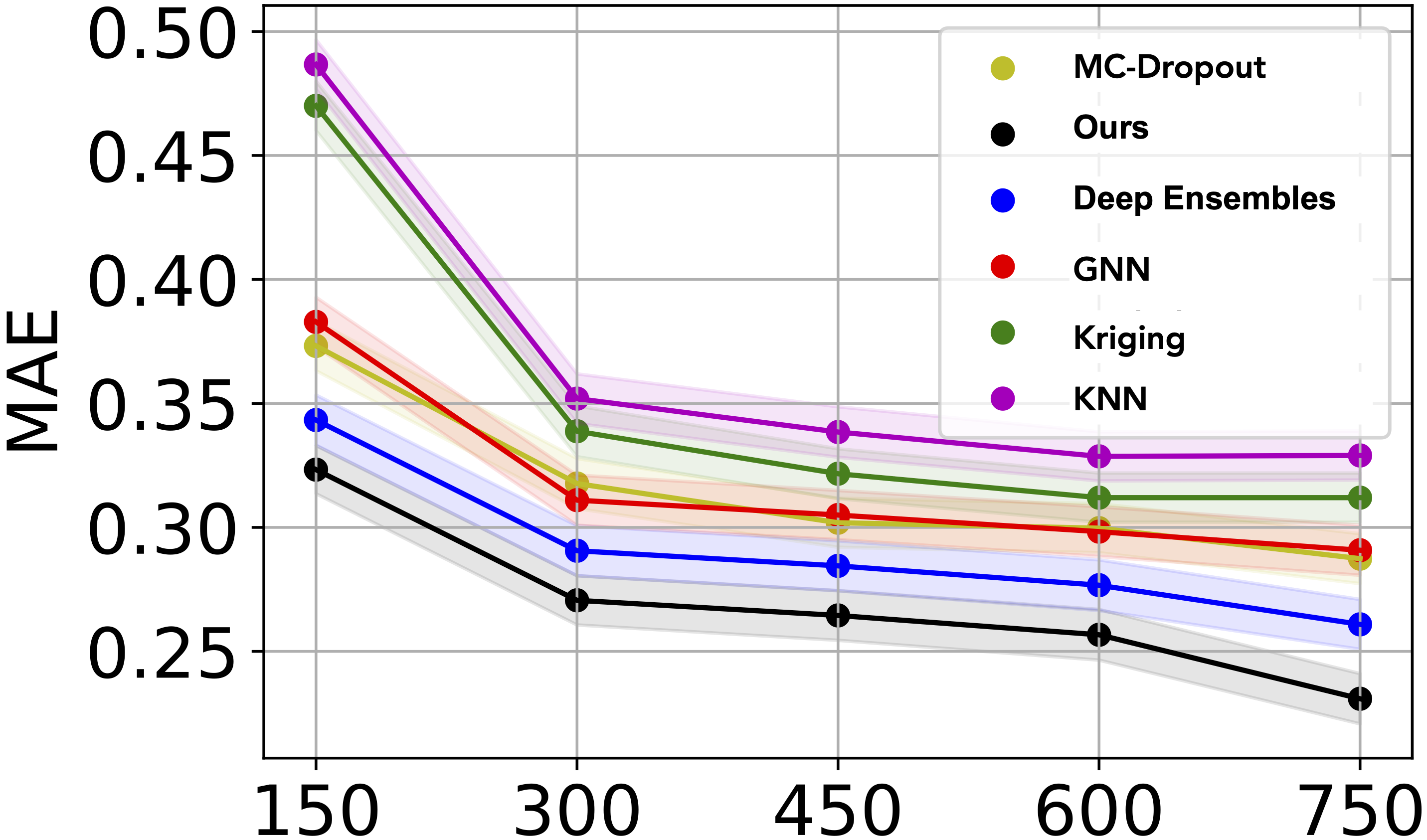}& 
        \hspace{-1mm}\includegraphics[width=0.24\textwidth]{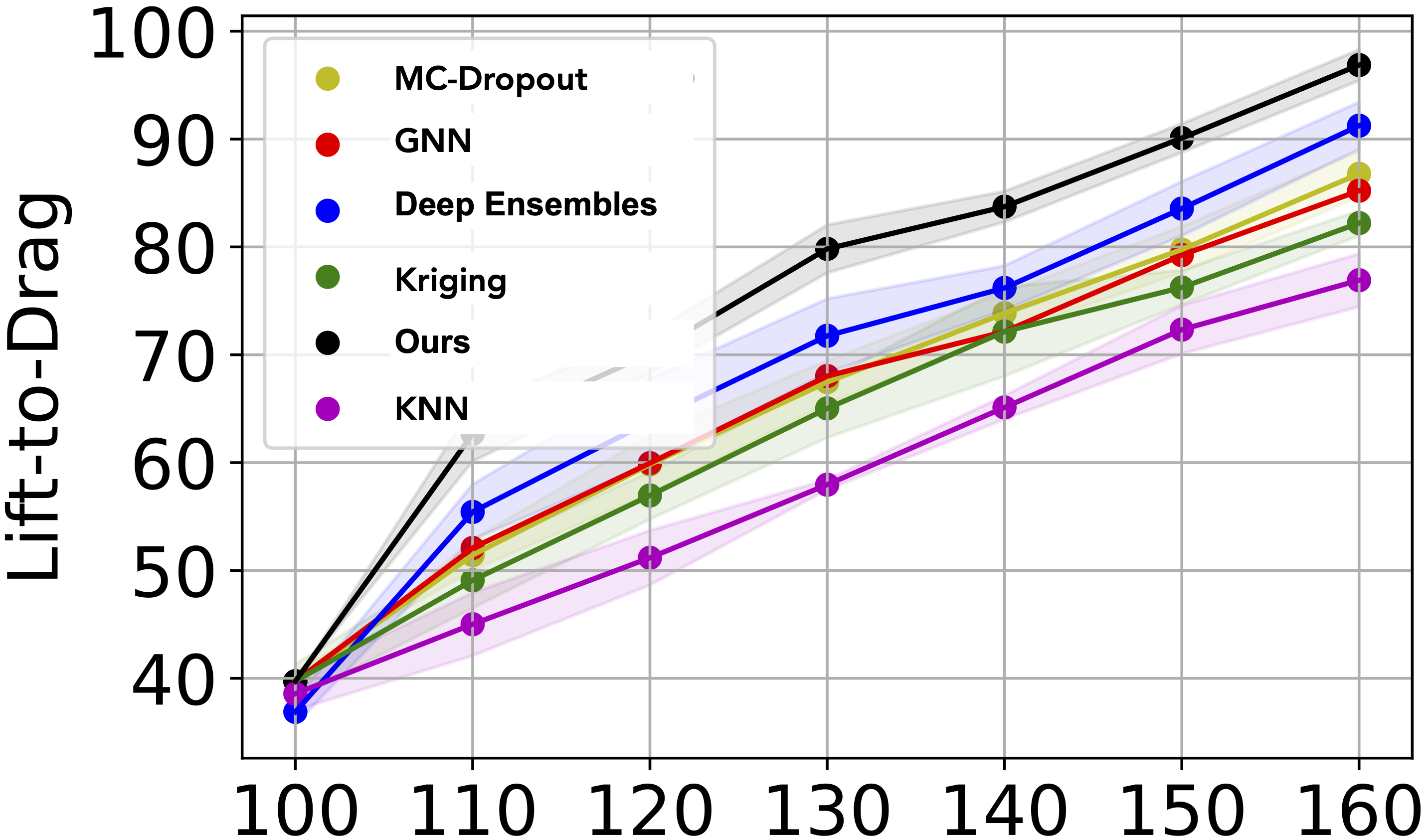} & \hspace{-3mm}\includegraphics[width=0.24\textwidth]{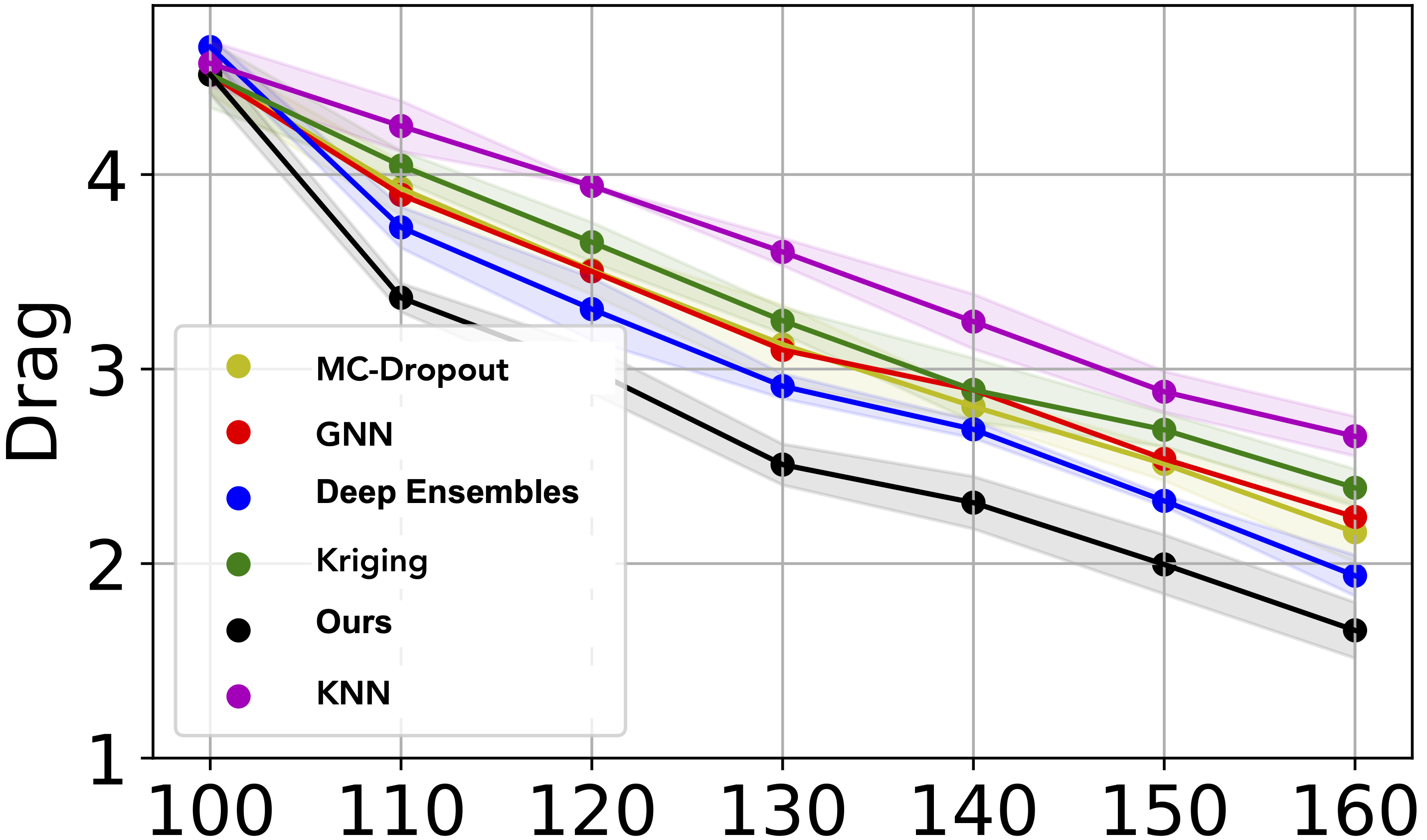} \\[-1mm]
        {\small (Airfoils)} & {\small (Cars)} & {\small (Airfoils)} & {\small (Cars)} 
    \end{tabular}   
    \vspace{-2mm} 
    \caption{\small 
       {\bf Left.}  Accuracy of the lift-to-drag estimate as a function of the number of exemplars used to train the emulators. {\bf Right.}  Lift-to-drag ratio of the shapes during optimization, as a function of number of iterations. }
      \label{fig:comp}
\end{figure*}

As all six methods being compared rely on an emulator,  the left {\it Airfoil} and {\it Cars} plots in Fig.~\ref{fig:comp} depict the accuracy of each on the test set as a function of the number of samples from the training set used to train it. \textit{Ours} outperforms the others consistently, especially when there are only a few training examples. For this accuracy evaluation, at each iteration, we add $100$ new samples for \textit{Airfoils} and $150$ for \textit{Cars}.

\begin{figure*}
    \centering
    \begin{tabular}{cc}
        \hspace{-2mm}\includegraphics[width=0.48\textwidth]{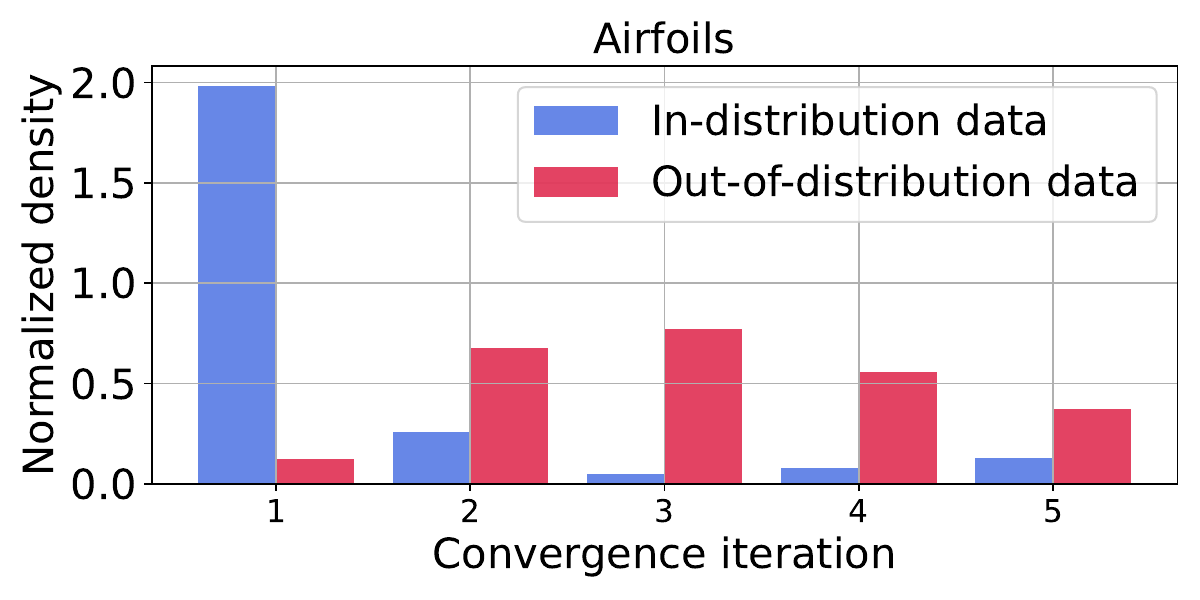}&
        \hspace{-2mm}\includegraphics[width=0.48\textwidth]{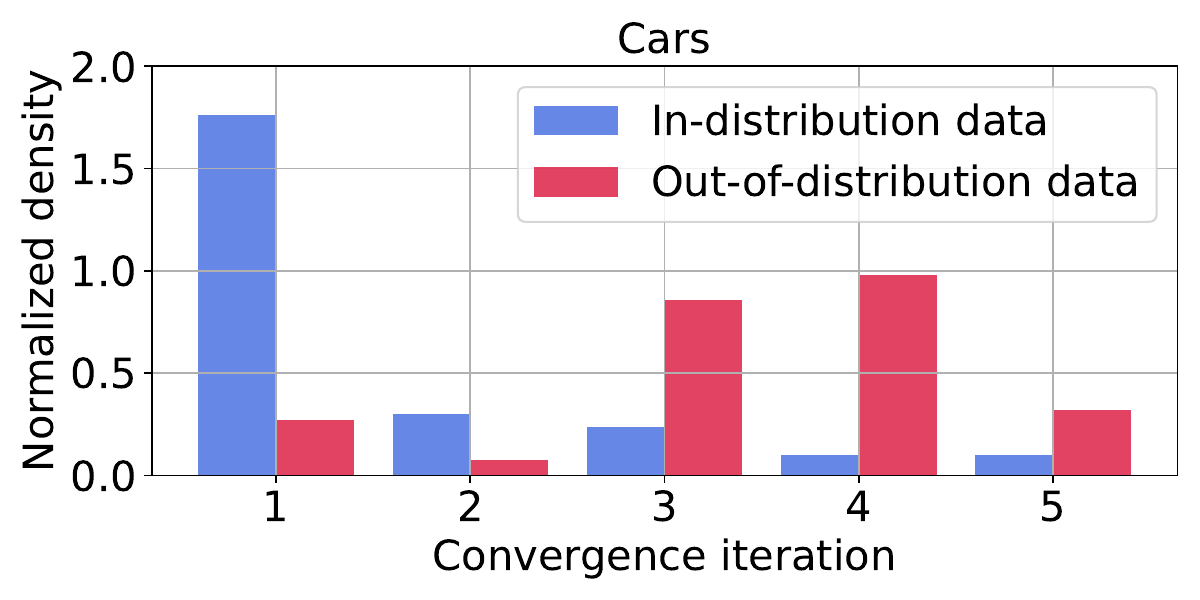}\\[1mm]
    \end{tabular}  
    \vspace{-6mm}   
    \caption{\small 
        \textbf{Convergence rates for in- and out-of-distribution samples.}  We plot the distribution of the number of iterations to convergence of our iterative GNNs for in-distribution vs. out-of-distribution samples from the test sets of airfoils and cars. In general, convergence takes significantly fewer steps for in-distribution samples than for out-of-distribution ones.}
    \label{fig:convHist}
\end{figure*}

As in the delineation experiments, we also evaluate the quality of uncertainty estimates through the lens of Out-of-distribution Detection~\cite{Fort21} task. To this end, given all the 3D shapes we have, we took the 200 top-performing ones in terms of their lift-to-drag ratio to be the out-of-distribution samples. For both datasets, the remaining shapes were then considered as the in-distribution ones. One thousand of these were used to train the emulators, and the others were used for testing purposes. After training, we generated uncertainty values for each shape in the in-distribution and out-of-distribution test sets. Finally, we computed standard ROC-AUC, PR-AUC~\citep{Malinin18} metrics for in- or out-of-distribution classification based on the uncertainty estimate. As can be seen in the top rows of Tab.~\ref{tab:aero_unc_results}, our approach generates uncertainty of a quality similar to that of ensembles. Furthermore, as shown in  Fig.~\ref{fig:convHist}, our method often only require 3 iterations to converge. This makes them a little faster than an ensemble of 5 ordinary GNNs and, importantly, requires far less memory and training time. We provide more details in Appendix~\ref{app:training}.

\begin{figure}[htb]
    \centering
        \begin{tabular}{c}
            \includegraphics[width=0.45\textwidth]{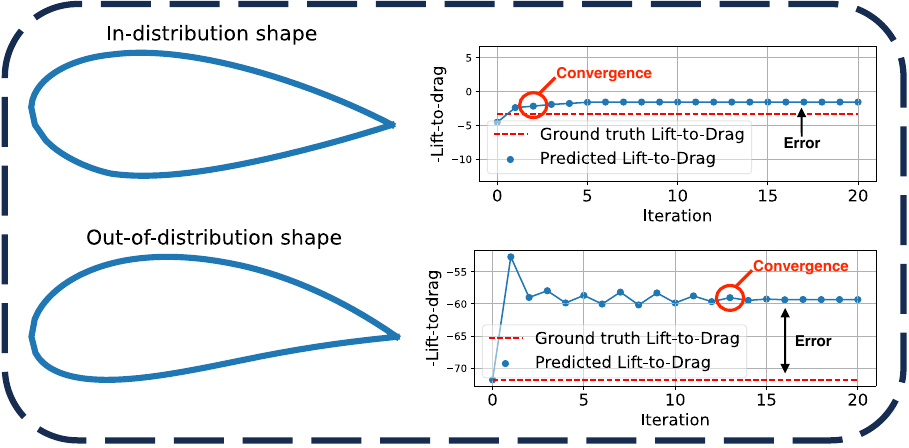} \\
        \end{tabular}%
    \caption{\small 
        \textbf{Convergence rate vs error.} For the in-distribution airfoil at the top, the consecutive values of predicted lift-to-drag values converge quickly and the limit is very close to the correct answer. By contrast, for an out-of-distribution airfoil below, the convergence is much slower and the limit is wrong. This is a behavior that we have consistently observed in our experiments.}
    \label{fig:conv_airfoils}
\end{figure}

We now turn to shape optimization using each one of the 6 methods. In each case, we used 100 randomly chosen samples from the training set, along with the corresponding simulations, to train the initial emulator. The rest of the training set, plus the OOD set, were treated as a set of unlabelled shapes. After the initial training, we ran the inference for each shape in it. For non-uncertainty approaches (\knn{} and \gnn{}), this yielded predicted performance values, and for the other values of the acquisition function (UCB~\cite{Auer02} with $\lambda = 3$). We sorted the unlabelled shapes according to these values and picked the 10 best. For GNN-based methods, for each one of these 10 shapes, we also performed 10 steps of gradient-based optimization~\cite{Kingma15a}. This relatively small number of iterations was chosen to allow us to reap the benefits of GNN-based shape optimization~\cite{Baque18}, without moving too far away from the starting points and producing shapes whose acquisition value is too different from that of the starting point. We discuss the influence of the number of iterations we perform in Appendix~\ref{app:gradient}. Finally, we ran simulations for these chosen shapes, added them to the training set, and iterated. For each method, we ran this whole process three times and plot the resulting lift-to-drag ratios as a function of the number of BO iterations performed in the {\it airfoil} plot in Fig.~\ref{fig:comp}. The shaded areas depict the corresponding variances. Again, our method outperforms the other approaches by a statistically significant margin.



Recall from Section~\ref{sec:method} that our approach is predicated on the fact that convergence of the iterative GNNs can be expected to be slower for out-of-distribution samples than for in-distribution ones. The plot on the left side of Fig.~\ref{fig:convHist} validates this hypothesis on the in-distribution and out-of-distribution splits.

\section{Ablation study}
\label{sec:ablations}

\subsection{Calibration Evaluation}

Though rAULC and correlation metrics from Sec.~\ref{sec:experiments} provide comprehensive information about calibration quality, we also provide the results for Expected Calibration Error (ECE)~\cite{Guo2017} evaluation. Tab.~\ref{tab:ece_results} provides ECE values for different methods, including MC-Dropout, Ensembles, and our proposed method. The results show that our method achieves better calibration than both Ensembles and MC-Dropout for all datasets.

\begin{table}[ht!]
    \centering
    \begin{tabular}{c|c|c|c}
    & \textit{MC-Dropout} & \textit{Ensembles} & \textit{Ours} \\
    \hline
    ECE (RT) & \cellcolor{secondbest}\graybold{0.997} & 1.138 & \cellcolor{best}\textbf{0.475} \\
    ECE (MS) & \cellcolor{secondbest}\graybold{0.558} & 0.794 & \cellcolor{best}\textbf{0.419} \\
    \hline
    ECE (Airfoils) & 1.758 & \cellcolor{secondbest}\graybold{1.162} & \cellcolor{best}\textbf{1.142} \\
    ECE (Cars) & 0.267 & \cellcolor{secondbest}\graybold{0.232} & \cellcolor{best}\textbf{0.227} \\
    \end{tabular}
    \caption{\small {\bf Expected Calibration Error (ECE) evaluation.} As it was previously demonstrated for the rAULC metric, our method outperforms other approaches in terms of ECE calibration.}
    \label{tab:ece_results}  
\end{table}

\subsection{Image Classification}

We expanded our experimental evaluation to include the widely-used task of image classification on the MNIST dataset, a popular benchmark for out-of-distribution (OOD) detection and uncertainty estimation. For the OOD task, we used FashionMNIST as the OOD samples, which is a standard choice in this context. To add more variability, we ran these experiments with two model architectures: 


\begin{itemize}
    \item \textbf{CNN Architecture:} A convolutional neural network (CNN) with several 2D convolutional layers followed by a fully-connected classification head.
    \item \textbf{MLP Architecture:} A multilayer perceptron (MLP) with 5 fully-connected layers, treating images as 784-dimensional vectors.
\end{itemize}

We evaluated these experiments using the same metrics as in our previous evaluations, ensuring consistency and comparability. The results are summarized in Table~\ref{tab:mnist_results}. As for our previous results, our method outperforms other approaches both in terms of model's accuracy and uncertainty quality.

\begin{table}[ht!]
    \centering
    \begin{tabular}{c|c|c|c|c}
    & \textit{MC-DP} & \textit{Ensembles} & \textit{Ours} & \\
    \hline
    Acc & 97.8 & \cellcolor{secondbest}\graybold{98.8} & \cellcolor{best}\textbf{99.0} & \multirow{6}{*}{\rotatebox{270}{CNN}} \\
    ECE & 0.021 & \cellcolor{secondbest}\graybold{0.018} & \cellcolor{best}\textbf{0.013} & \\
    ROC-AUC & 94.4 & \cellcolor{best}\textbf{98.5} & \cellcolor{best}\textbf{98.5} & \\
    PR-AUC & 93.8 & \cellcolor{secondbest}\graybold{98.1} & \cellcolor{best}\textbf{98.5} & \\
    Train & \cellcolor{best}1x & 5x & \cellcolor{secondbest}\graybold{3x} & \\
    Inf & \cellcolor{secondbest}\graybold{5x} & \cellcolor{secondbest}\graybold{5x} & \cellcolor{best}3x & \\
    \hline
    Acc & 96.1 & \cellcolor{best}\textbf{97.7} & \cellcolor{secondbest}\graybold{97.4} & \multirow{6}{*}{\rotatebox{270}{MLP}} \\
    ECE & 0.026 & \cellcolor{secondbest}\graybold{0.022} & \cellcolor{best}\textbf{0.020} & \\
    ROC-AUC & 61.5 & \cellcolor{secondbest}\graybold{89.5} & \cellcolor{best}\textbf{89.6} & \\
    PR-AUC & 69.2 & \cellcolor{best}\textbf{89.9} & \cellcolor{secondbest}\graybold{88.7} & \\
    Train & \cellcolor{best}1x & 5x & \cellcolor{secondbest}\graybold{3x} & \\
    Inf & \cellcolor{secondbest}\graybold{5x} & \cellcolor{secondbest}\graybold{5x} & \cellcolor{best}\textbf{3x} & \\
    \end{tabular}
    \caption{\small {\bf  MNIST classification results for CNN (top), and MLP (bottom) architectures.} The best result in each category is in \textbf{bold} and the second best is in \graybold{bold}. Most correspond to Ours and Deep Ensembles.}
    \label{tab:mnist_results}
\end{table}


\section{Conclusion}

We have presented an approach to assessing the quality of predictions by iterative networks at a much lower cost than Deep Ensembles, currently the most reliable approach to such an assessment, and more reliably than other state-of-the-art methods. Our method relies on measuring how fast successive estimates converge and does require neither any change in network architecture nor training more than one. In the shape optimization part of this work, we have focused on aerodynamics but the principle applies to many other devices, ranging from the cooling plates of an electric vehicle battery to the optics of an image acquisition device. In future work, we will therefore explore a broader set of potential applications. 

\clearpage
\section*{Impact Statement}
This paper presents work whose goal is to advance the field of Machine Learning. There are many potential societal consequences of our work, none which we feel must be specifically highlighted here.

\bibliographystyle{icml2024}
\bibliography{./bib/string,./bib/vision,./bib/learning,./bib/cfd,./bib/misc,./bib/stats,./bib/optim,./bib/graphics,./bib/biomed,./new}


\newpage
\appendix
\twocolumn
\section{Appendix}

In this section, we first examine the behavior of iterative networks in a very simple case. We then provide details about the training procedure and additional supporting evidence for some of the claims made in the paper.
	
\subsection{Analysis of a Simple Case}
\label{app:analysis}



\begin{figure*}[htb]
\begin{tabular}{ccc}
   \includegraphics[height=3.2cm]{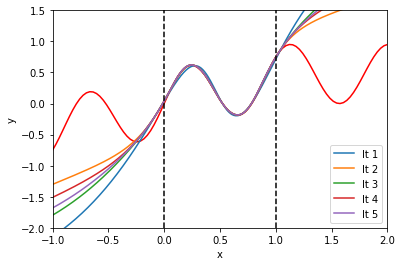}&
   \includegraphics[height=3.2cm]{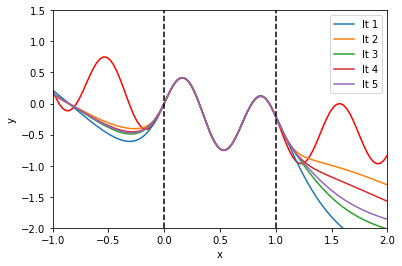}&
   \includegraphics[height=3.2cm]{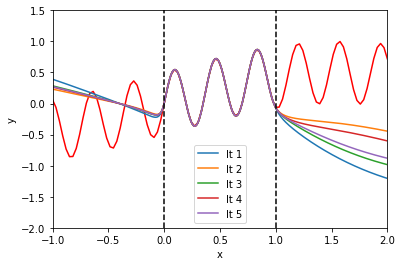}\\
   \includegraphics[height=3.2cm]{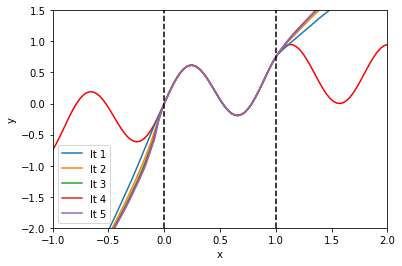}&
   \includegraphics[height=3.2cm]{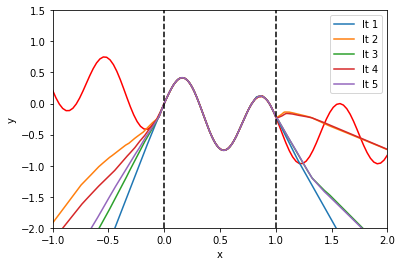}&
   \includegraphics[height=3.2cm]{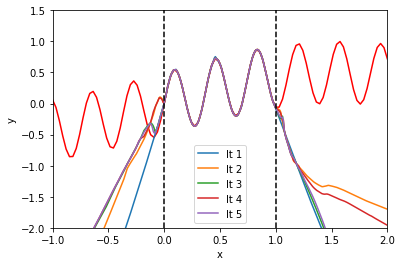}\\
   (a = 3, b = 4) &  (a = 5, b = 4) &  (a = 8, b = 9)
\end{tabular}
    \vspace{-2mm}
    \centering
    \caption{\small 
    \textbf{Learning to interpolate a 1D function}. Using a two-layer perceptron to interpolate $f(x) = \sin(a *x) \cos (b *y)$ given training pairs $(x,f(x))$ for which $0 < x < 1$. Each curve represents the value of $y_i^t$ from Eq.~\ref{eq:interpol1} for values of $x$ ranging from -1.0 to 2.0, that is, both inside and outside the training domain. There is one curve per iteration $i$ in Eq.~\ref{eq:interpol1}, ranging from 1 to 5.  {\bf Top row.} Taking $\tanh$ to be the activation function. {\bf Bottom row.} Using ReLu. }
    \label{fig:interpol1}
\end{figure*}

To model the behavior of our iterative networks in a simpler and easier-to-analyze context, we replace CNNs and GNNs with a perceptron $f_W$ that takes two scalar inputs $x$ and $y$ and outputs a scalar. Given a training set $\{(x_i,r_i) \; , 1 \leq i \leq N \}$, we make it iterative by computing
\begin{align}
y_i^1 &= f_{W} (x,0)  \nonumber \\
y_i^2 &= f_{W} (x,y_i^1) \nonumber \\
... \nonumber \\
y_i^t  &= f_{W}(x,y_i^{t-1}) \label{eq:interpol1}
\end{align}
%
for each $i$, where $t_i$ is a different random integer between 1 and $T$ for each sample. In these examples, we use $T=5$. We then minimize the total loss $\sum_i (r_i -  y_i^{(t_i)})^2$.  Fig.~\ref{fig:interpol1} depicts the results of this process when the $x_i$ are uniformly sampled between 0 and 1 and the $r_i$ are taken to be $sin(a*x_i)*cos(b*x_i)$ for different values of $a$ and $b$. For values of $x$ between 0 and 1, that is, for values that are within the training domain, we have $y_i^0 \approx y_i^1 .... \approx  y_i^T \approx r_i$. In contrast, out of domain, that is, outside the range [0,1], this is not true anymore, and we can see strong oscillations of the successive $y_i^t$ values for $1 \leq t \leq T$. This makes sense because deep networks are known not to extrapolate well. Thus, even though the network is trained to produce similar predictions for all values of $t$ in-domain,  the out-of-domain predictions are essentially random, and there is no reason for them to be equal. In the results section, we showed that, for both airfoils and car shapes, out-of-domain values of $x$ tend to produce oscillations and slow convergence. Interestingly, we observe exactly the same behavior on this very simple example, as evidenced by the fact that the curves of Fig.~\ref{fig:interpol1} are {\it not} superposed for $x<0$ and $x>1$.


\begin{figure*}[htb]
\begin{tabular}{cc}
   \includegraphics[height=4.5cm]{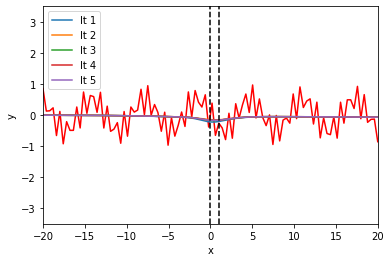}& \includegraphics[height=4.5cm]{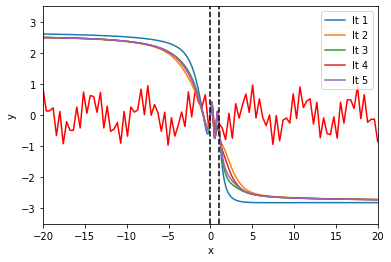}\\
   \includegraphics[height=4.5cm]{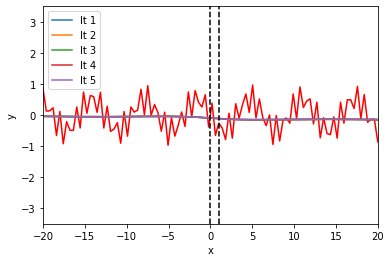}& \includegraphics[height=4.5cm]{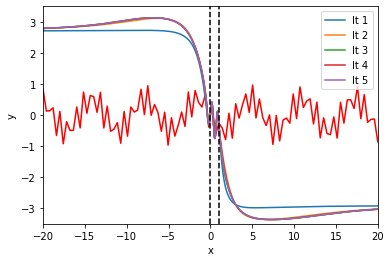}\\
   (Before training) & (After training)
\end{tabular}
    \centering
    \caption{\small 
    \textbf{Influence of initialization.} We plot the same curves as in Fig.~\ref{fig:interpol1} when learning to interpolate the function $f(x) = \sin(5 *x) \cos (4 *y)$, but over a more extended range of $x$ and starting from a different initialization of the perceptron weights in each row. 
    {\bf Before training.}  As before, each curve represents the values $y_i^t$  as a function of $x$. Here we plot those returned by our perceptrons after initialization of their activation weights,  but before actual training. The two plots correspond to the two different initializations. {\bf After training.}  Values after training. There are similar for $0 < x < 1$ but different out of this domain. Not that they are also very different across the two rows because of the slightly different initializations.}
    \label{fig:interpol2}
\end{figure*}

The exact values obtained for these out-of-domain samples are very hard to predict. As can be seen by comparing the two rows of Fig.~\ref{fig:interpol1}, they depend critically on the chosen activation function, $\tanh$ or ReLu in this case. They also depend heavily on how the networks have been initialized, as can be seen in Fig.~\ref{fig:interpol2}. In one case, we initialized the weights of our perceptrons using normally distributed weights. In the other, we used the slightly more sophisticated Xavier Initialization~\citep{Kumar17b}.  

Crucially, in all cases, seeing large variations in the values of the successive $y_k(x)$ for a given $x$ is {\it always} a warning sign that the estimated value is likely to be incorrect. This is what we exploit in this work.

\subsection{Bayesian Optimization}
\label{sec:bo}

Given a performance estimator of uknown reliability, exploration-and-exploitation techniques seek to find global optimum of that estimator while at the same time accounting for potential inaccuracies in its predictions. 

Bayesian Optimization (BO)~\citep{Mockus12} is one of the best-known approaches to finding global minima of a black-box function $g: \mathbf{{A}}  \rightarrow \mathbb{R}$, where $\mathbf{{A}}$ represents the space of possible shapes, without assuming any specific functional form for $g$. It is often preferred to more direct approaches, such as the adjoint method~\citep{Allaire15}, when $g$ is expensive to evaluate, which often is the case when $g$ is implemented by a physics-based simulator. 

\begin{figure}[htbp]
    \centering
    \includegraphics[width=0.43\textwidth]{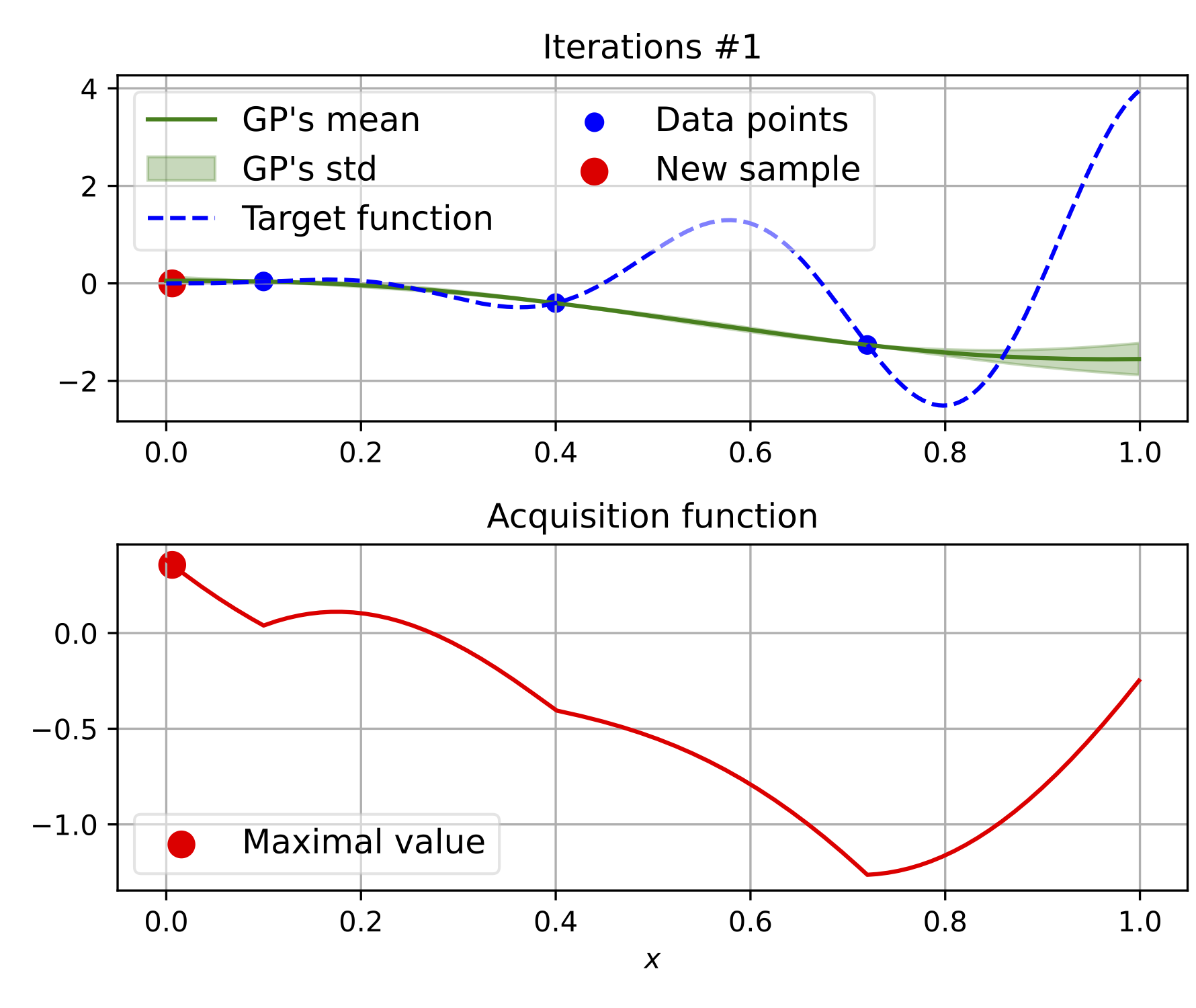}
    \includegraphics[width=0.43\textwidth]{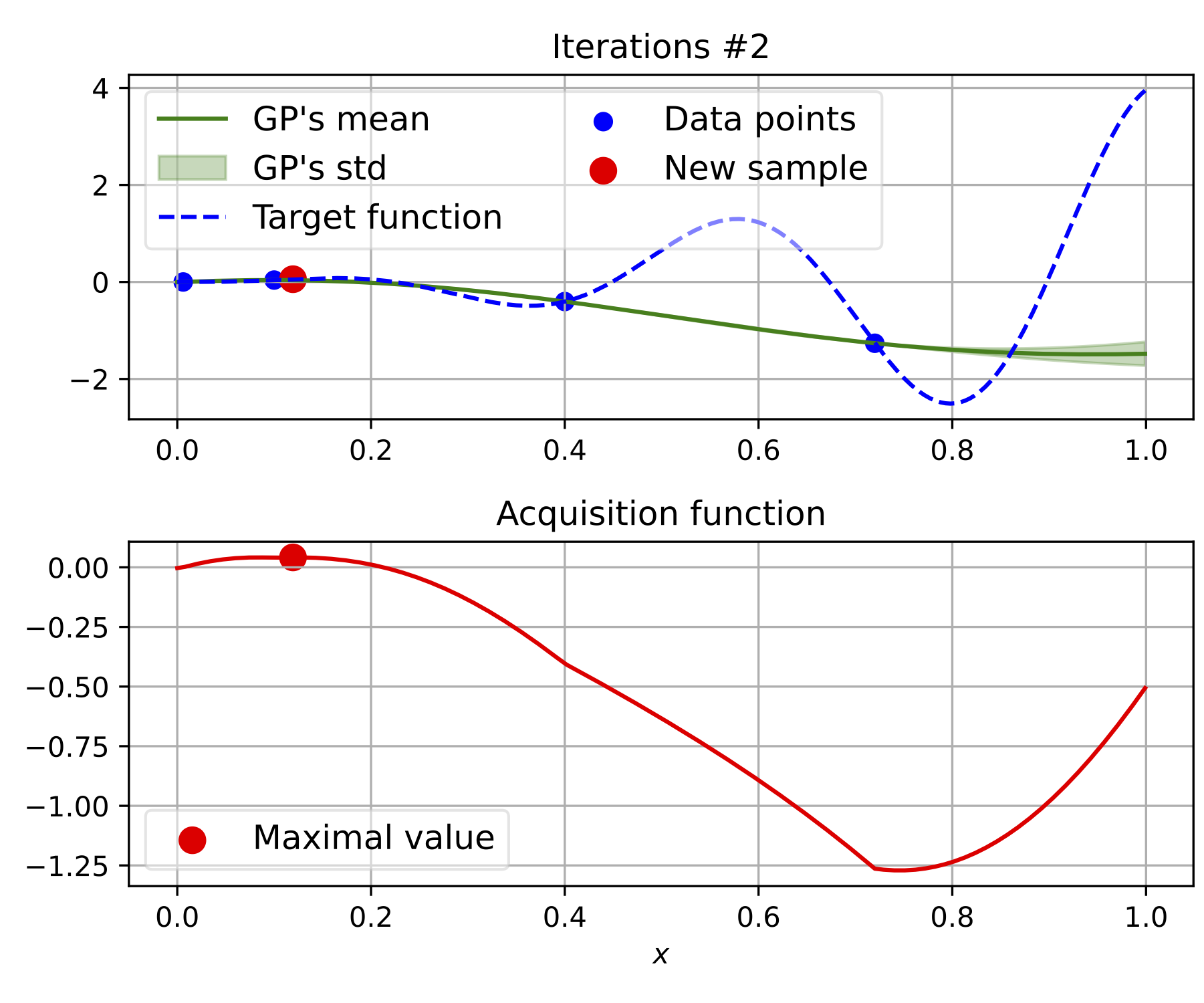}
    \includegraphics[width=0.43\textwidth]{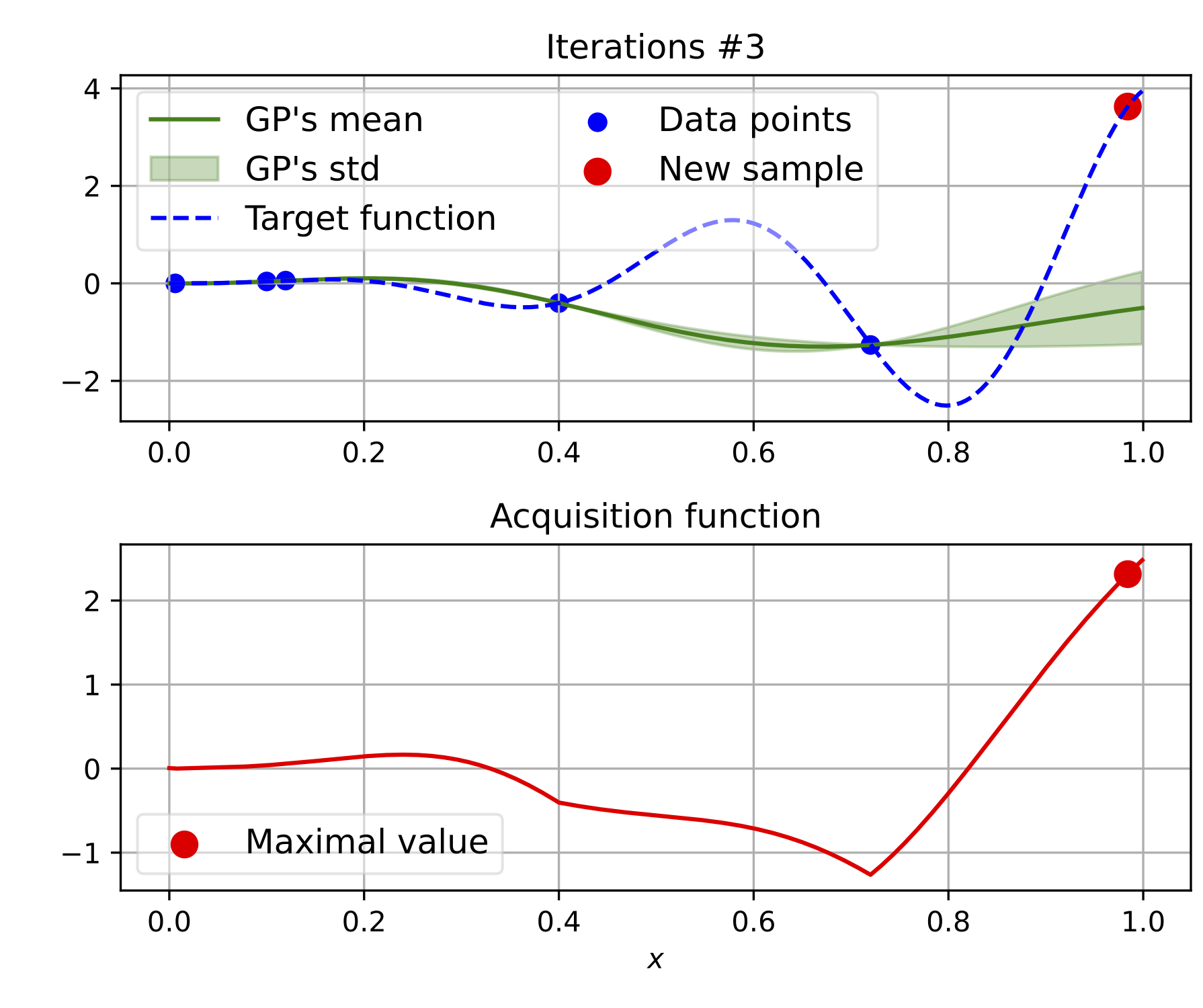}
    \caption{\textbf{Bayesian Optimization.} Given three initial data points for the function (dashed blue), we want to optimize, we train a GP surrogate model (Step 1) and compute the UCB acquisition function~\citep{Auer02} over the $[0, 1]$ range (Steps 2-3). We then select the points that maximize, evaluate the target function at those points, and include the results in our training dataset (Step 4). The process is then iterated and, eventually, we find the true maximum of the function at $x \approx 1$, whereas a simple gradient based method would probably have remained trapped at the local maximum $x \approx 0.58$.}
    \label{fig:bo_example}
\end{figure}

BO typically starts with a surrogate model $f_{\Theta}: \textbf{A}  \rightarrow \mathbb{R}$ whose output depends on a set of parameters $\Theta$. $f_{\Theta}$ is assumed to approximate $g$, to be fast to compute, and to be able to evaluate the reliability of its own predictions in terms of a uncertainty. It is used to explore $\mathbf{A}$ quickly in search of a solution of $\textbf{x}^{*} = \operatornamewithlimits{argmin}_{\textbf{x} \in \mathbf{{A}}} g(\mathbf{x})$. Given an initial training set $\{(\textbf{x}_{i}, r_{i})\}_{i}$ of input shapes $\textbf{x}_{i}$ and outputs $r_{i} = g(\textbf{x}_{i})$, it iterates the following steps:
\begin{tcolorbox}[colback=white, colframe=black, sharp corners, boxrule=1pt, boxsep=1pt]
\begin{itemize} \label{itemize:steps}
 \setlength{\itemindent}{-7mm}
 \item[] \label{step1} \textbf{Step 1:} Find $\Theta$ that yields the best possible prediction by $f_{\Theta}$.
 
 \item[] \textbf{Step 2:} Generate new samples not present in the training set.
 
 \item[] \textbf{Step 3:} Evaluate an \textit{acquisition function} on these samples.
 
 \item[] \textbf{Step 4:} Add the best ones to the training set and go back to Step 1.
\end{itemize}
\end{tcolorbox}
As shown in the example of Fig.~\ref{fig:bo_example}, the role of the acquisition function is to gauge how desirable it is to evaluate a point, based on the current state of the model. It is often  taken to be the  \textit{Expected Improvement}  (EI)~\citep{Qin17} or \textit{Upper Confidence Bound} (UCB)~\citep{Auer02} that favor samples with the greatest potential for improvement over the current optimum. It is computed as a function of the values predicted by the surrogate and their associated uncertainty.

\subsection{Training setups}
\label{app:training}

For our experiments, we used single Tesla V100 GPU with 32Gb of memory. The training process was implemented using the Pytorch~\citep{Paszke17} and Pytorch Geometrics~\citep{Fey19} frameworks.


\begin{table*}
    \centering
    \begin{tabular}{c|cccc|c} 
    &  \textit{GNN} & \textit{Deep Ensembles} & \textit{MC-Dropout} & \textit{Ours} & \\
    \hline
    Memory & 1x &  5x & 1x & 1x & \multirow{3}{*}{\rotatebox[origin=c]{270}{AIR}} \\ 
    Inf. Time & 1x & 5x & 5x & 3x & \\  
    Train. Time & 1x & 5x & 1x & 2x & \\  
    \hline
    Memory & 1x & 5x & 1x & 1x & \multirow{3}{*}{\rotatebox[origin=c]{270}{CAR}} \\ 
    Inf. Time & 1x & 5x & 5x & 3x & \\
    Train. Time & 1x & 5x & 1x & 2x & \\
    \end{tabular}
    \caption{\small {\bf Computational costs.} The Memory, Inference Time, and Training Time metrics measure the amount of time and memory required to train the network(s) and to perform inference, in comparison to a single model.}
    \label{tab:comp_costs}  
\end{table*}

\paragraph{Airfoils.} For airfoils, we have generated 1500 shapes from NACA parameters, and simulated pressure and lift-to-drag values with XFOIL simulator. As an emulator, we use architecture that consists of 35 GMM layers~\citep{Monti17} with ReLU activations. First, we extract node features with these GMM layers and pass them to pressure branch, that consists out of 3 GMM layers, and lift-to-drag branch, that uses global pooling and 3 fully-connected layers to predict final scalar. For training, we use Adam optimizer~\citep{Kingma15a} and perform 200 epochs with 128 batch size and $0.001$ learning rate. Both for lift-to-drag and pressure, we use mean squared error (MSE) loss and combine them into final loss with weights 1 for scalar and 100 for pressure.

\paragraph{Cars.} For cars dataset, we have generated 1500 shapes from MeshSDF vectors, and simulated pressure and drag values with OpenFOAM simulator. As an emulator, we use architecture that consists of 50 GMM layers with ELU activations~\citep{Clevert15} and skip-connections~\citep{He16a}. Similar to airfoils, we extract node features with these GMM layers and pass them to pressure branch, that consists out of 5 GMM layers, and drag branch, that uses global pooling and 5 fully-connected layers to predict final scalar. For training, we use Adam optimizer and perform 6 epochs with 8 batch size and $0.001$ learning rate. Both for lift-to-drag and pressure, we use mean squared error (MSE) loss and combine them into final loss with weights 1 for scalar and $1/200$ for pressure.

\subsection{Propagating Information} 
\label{app:propagate}

\begin{figure}
    \centering
    \begin{tabular}{cc}
        \includegraphics[width=0.22\textwidth]{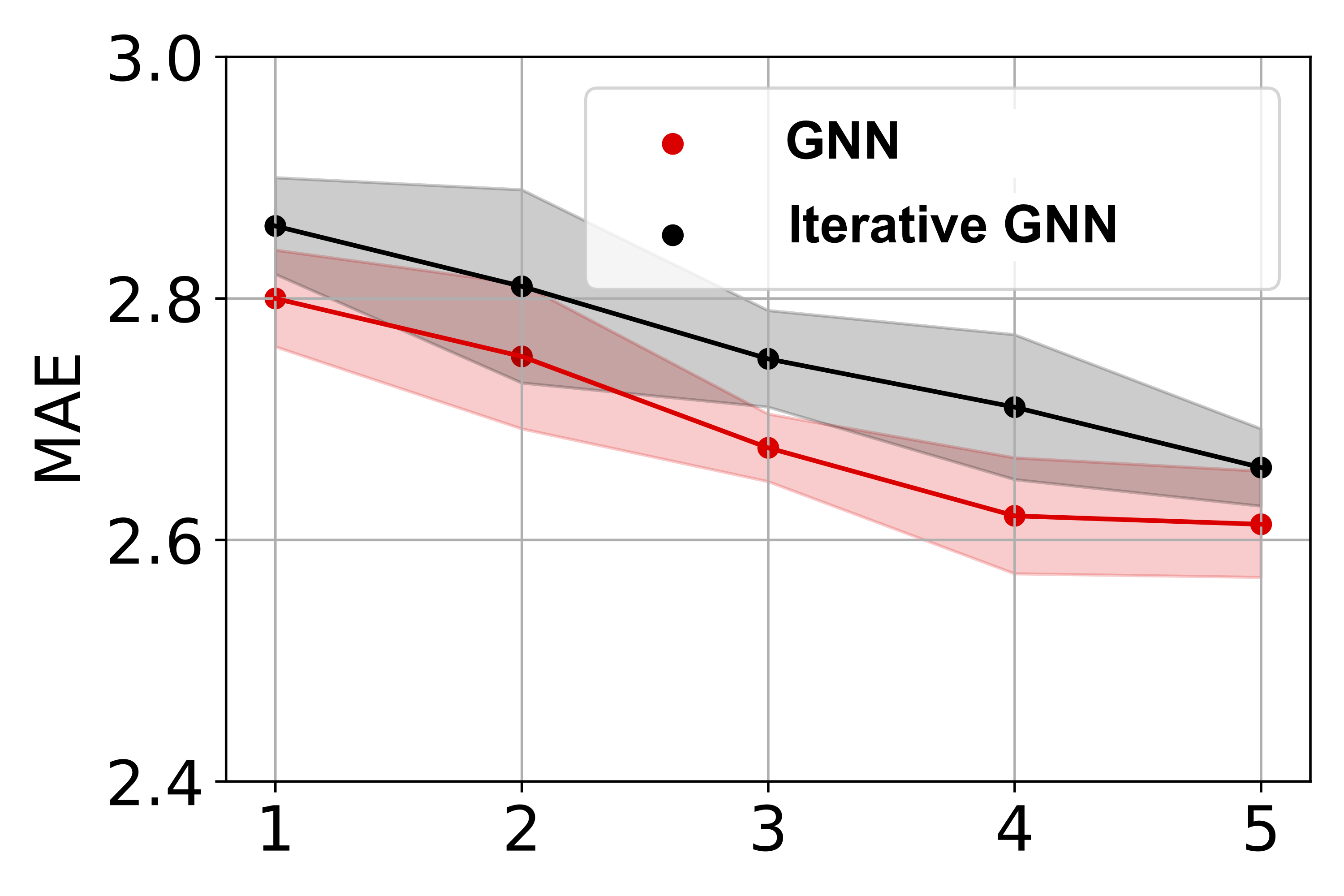}&
        \includegraphics[width=0.22\textwidth]{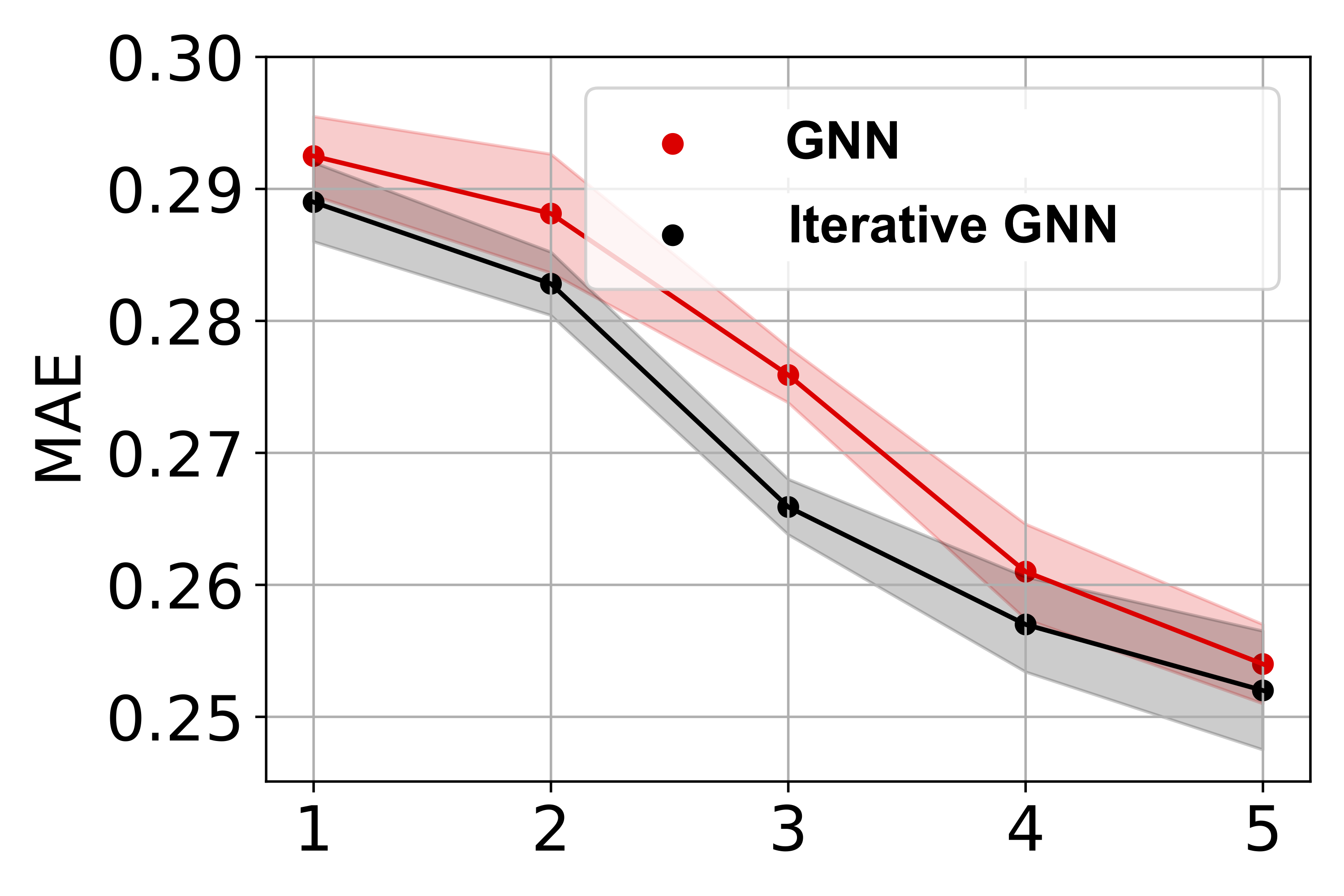}\\
        Lift-to-Drag & Drag
    \end{tabular}     
    \caption{\small 
        \textbf{Propagating information across an shape}. A comparable behavior is observed when increasing the depth of a standard GNN (red curves) and when running several iterations of a shallower iterative GNN (black curves).
        }
    \label{fig:receptive}
    \end{figure}

In a standard GNN information is propagated across the shape with each successive convolution. Hence, it is comparatively slow and our reentrant GNNs address this. To support, this claim we ran an experiment to test the influence of the receptive fields of the GNNs, which control the speed at which information percolates across the network. We trained $5$ airfoils and car emulator models of increasing depth while keeping total weights number fixed. Starting from the original architecture, we plot the prediction mean error for both lift-to-drag and drag in Fig.~\ref{fig:receptive} in red. As expected, the error decreases as depth increases and more information is propagated across the shape. The exact same behavior can be observed when using our GNN run iteratively, as shown by the black curves. This supports our claim that each iteration helps propagate the information across the shape just as effectively as when using the deeper network. 


\subsection{Gradient Optimization}
\label{app:gradient}


\begin{figure}
    \includegraphics[width=0.45\textwidth]{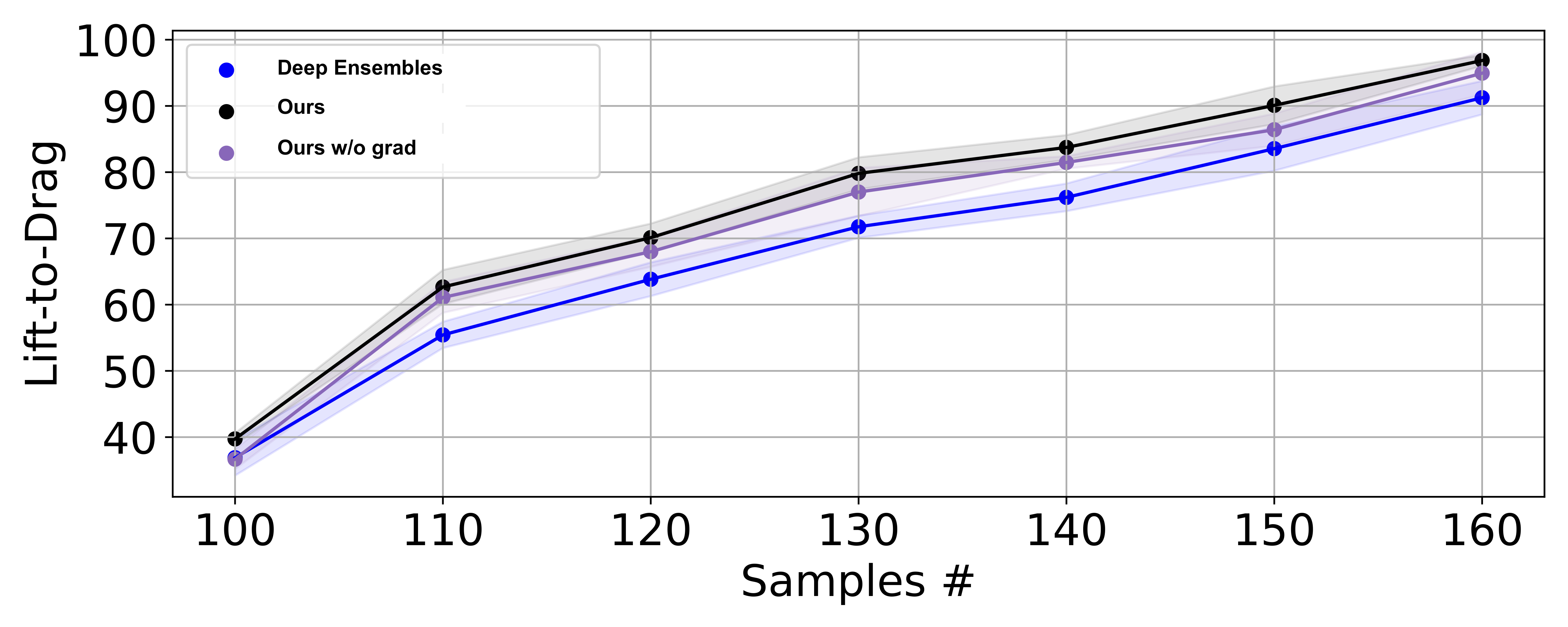}
    \vspace{-5mm}
    \centering
    \caption{\small 
    \textbf{Impact of refining the shapes.} Turning on gradient optimization of new samples delivers a small performance increase, but smaller than the one used by replacing ensembles with a version of our approach without the refinement.} 
    \label{fig:no_grad}
\end{figure}

Given the shapes selected according to the acquisition function during Bayesian Optimization, our method performs several gradient steps in order to refine these shapes and makes them more performant. In this subsection, we examine the impact of performing this optimization.  

In the results shown in the main paper, given the current state of the emulator, we performed 10 steps of an Adam-based optimizer with a $1e-4$ learning rate to refine each selected shape. In Fig.~\ref{fig:no_grad}, we plot the results obtained for the airfoils by doing this refinement (\textit{Ours}), not doing it (\textit{Ours} w/o grad), or using deep ensembles (\textit{Deep Ensembles}) baseline. \textit{Ours} without refinement  already delivers an improvement overs ensembles, with a further but smaller improvement when performing the refinement. We tried increasing the number of refinement steps but that brought no further improvement.


\newlength{\imgheight}
\setlength{\imgheight}{0.15\textwidth}
\begin{figure*}[!h]
	\centering
	\begin{tabular}{@{}>{\centering\arraybackslash}m{0.02\textwidth} 
			@{}>{\centering\arraybackslash}m{0.2\textwidth} 
			@{}>{\centering\arraybackslash}m{0.2\textwidth}
			@{}>{\centering\arraybackslash}m{0.2\textwidth} 
			@{}>{\centering\arraybackslash}m{0.2\textwidth} @{}}
		& \multicolumn{2}{c}{RoadTracer} & \multicolumn{2}{c}{Massachusetts}  \\
		\rotatebox[origin=l]{90}{\footnotesize Image} &
		\includegraphics[height=\imgheight]{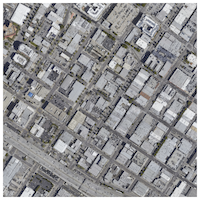} &
		\includegraphics[height=\imgheight]{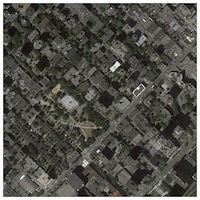} &
		\includegraphics[height=\imgheight]{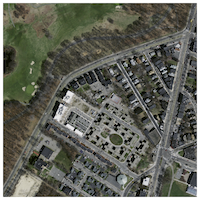} &
		\includegraphics[height=\imgheight]{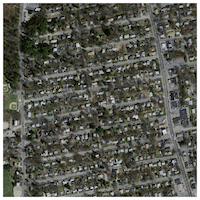} \\
		
		{\rotatebox[origin=l]{90}{\footnotesize Label }} &
		\includegraphics[height=\imgheight]{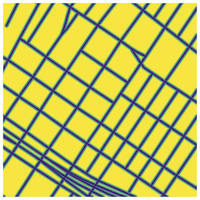} &
		\includegraphics[height=\imgheight]{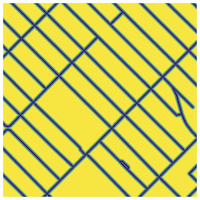} &
		\includegraphics[height=\imgheight]{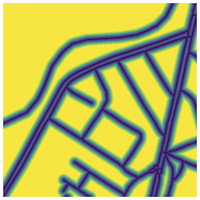} &
		\includegraphics[height=\imgheight]{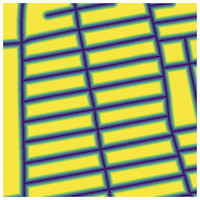} \\
		
		\rotatebox[origin=l]{90}{\footnotesize MC Pred. } &
		\includegraphics[height=\imgheight]{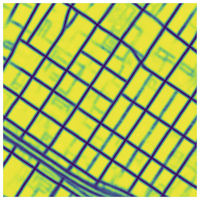} &
		\includegraphics[height=\imgheight]{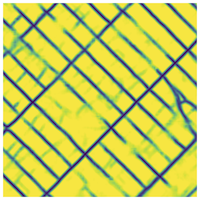} &
		\includegraphics[height=\imgheight]{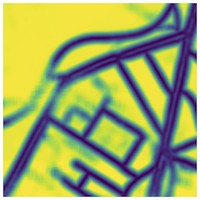} &
		\includegraphics[height=\imgheight]{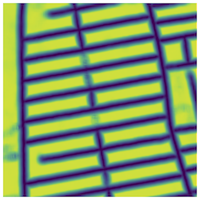} \\
		
		\rotatebox[origin=l]{90}{\footnotesize MC Unc. } &
		\includegraphics[height=\imgheight]{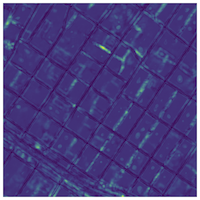} &
		\includegraphics[height=\imgheight]{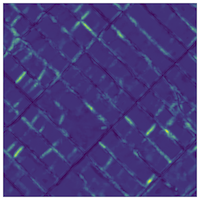} &
		\includegraphics[height=\imgheight]{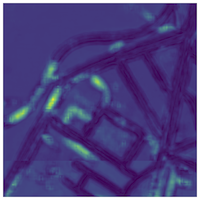} &
		\includegraphics[height=\imgheight]{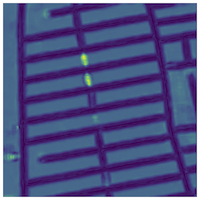} \\
		
		\rotatebox[origin=l]{90}{\footnotesize Ensemble Pred. } &
		\includegraphics[height=\imgheight]{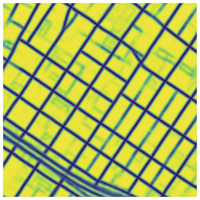} &
		\includegraphics[height=\imgheight]{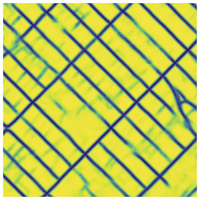} &
		\includegraphics[height=\imgheight]{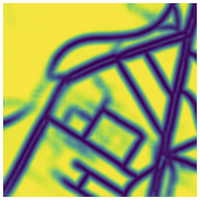} &
		\includegraphics[height=\imgheight]{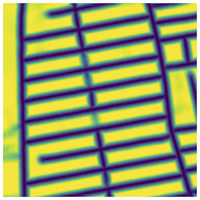} \\
		
		{\rotatebox[origin=l]{90}{\footnotesize Ensemble Unc. }} &
		\includegraphics[height=\imgheight]{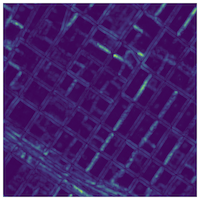} &
		\includegraphics[height=\imgheight]{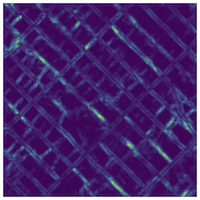} &
		\includegraphics[height=\imgheight]{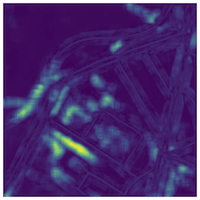} &
		\includegraphics[height=\imgheight]{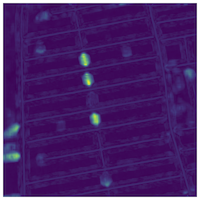} \\
		
		\rotatebox[origin=l]{90}{\footnotesize Ours Pred. } &
		\includegraphics[height=\imgheight]{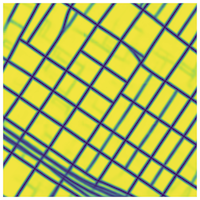} &
		\includegraphics[height=\imgheight]{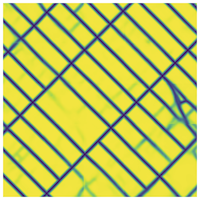} &
		\includegraphics[height=\imgheight]{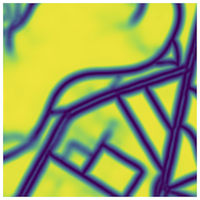} &
		\includegraphics[height=\imgheight]{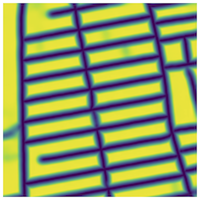} \\
		
		\rotatebox[origin=l]{90}{\footnotesize Ours Unc. } &
		\includegraphics[height=\imgheight]{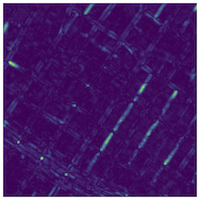} &
		\includegraphics[height=\imgheight]{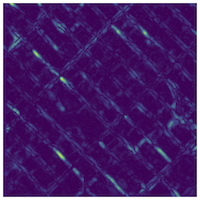} &
		\includegraphics[height=\imgheight]{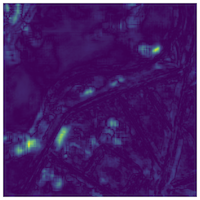} &
		\includegraphics[height=\imgheight]{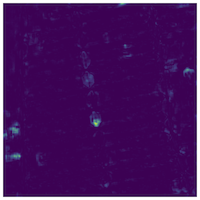} \\

	\end{tabular}
	\caption{ \label{fig:visual_combined}
		Test predictions and uncertainties produced by different methods on two datasets. 
	}
\end{figure*}

\clearpage
\subsection{Ensembles on Iterative Architecture}
\label{app:iter_ens}

Another baseline for comparison could involve applying Deep Ensembles (DE) to our iterative architecture. While this approach would significantly increase computational complexity, it could provide better metrics for uncertainty estimation. In this subsection, we compare the proposed DE baseline against our model.

In the Tables~\ref{tab:rt_dataset_results} and~\ref{tab:ms_dataset_results} below, we present the results of this comparison. While the DE baseline delivers good accuracy and uncertainty quality, it is approximately 14 times slower than the single model baseline and about 5 times slower than our approach. These results confirm that while DE applied to the iterative architecture provides high-quality uncertainty estimates and accuracy, the computational cost is significantly higher compared to our approach. Our method strikes a balance between computational efficiency and performance, making it a more practical choice for real-world applications where computational resources and time are critical factors.

\begin{table}[tp]
    \centering
    \begin{tabular}{c|c|c|c|c}
    & \textit{MC-DP} & \textit{Ens.} & \textit{Ens. Iter.} & \textit{Ours} \\
    \hline
    Corr & 87.1 & \cellcolor{secondbest}\graybold{87.4} & \cellcolor{best}\textbf{88.1} & 85.2 \\
    Comp & 58.2 & 66.7 & \cellcolor{secondbest}\graybold{76.1} & \cellcolor{best}\textbf{77.8} \\
    Qual & 54.1 & 60.8 & \cellcolor{best}\textbf{69.1} & \cellcolor{secondbest}\graybold{68.6} \\
    F1 & 20.4 & 22.1 & \cellcolor{secondbest}\graybold{24.4} & \cellcolor{best}\textbf{24.5} \\
    APLS & 58.78 & 68.81 & \cellcolor{best}\textbf{78.40} & \cellcolor{secondbest}\graybold{77.21} \\
    \hline
    rAULC & 30.18 & \cellcolor{best}\textbf{72.19} & 69.03 & \cellcolor{secondbest}\graybold{69.23} \\
    Corr (unc.) & 59.72 & \cellcolor{best}\textbf{79.42} & 79.17 & \cellcolor{secondbest}\graybold{74.73} \\
    ECE & 0.997 & 1.138 & \cellcolor{secondbest}\graybold{0.850} & \cellcolor{best}\textbf{0.475} \\
    \hline
    Train & \cellcolor{best}1x & 5x & 14x & \cellcolor{secondbest}\graybold{2.8x} \\
    Inf & \cellcolor{secondbest}\graybold{5x} & \cellcolor{secondbest}\graybold{5x} & 13.5x & \cellcolor{best}2.7x \\
    \end{tabular}
    \caption{\small {\bf Uncertainty and performance metrics on RT Dataset.} The best result in each category is in \textbf{bold} and the second best is in \graybold{bold}. Most correspond to Ours and Iterative Ensembles.}
    \label{tab:rt_dataset_results}
\end{table}

\begin{table}[!hbp]
    \centering
    \begin{tabular}{c|c|c|c|c}
    & \textit{MC-DP} & \textit{Ens.} & \textit{Ens. Iter.} & \textit{Ours} \\
    \hline
    Corr & 81.6 & 83.6 & \cellcolor{best}\textbf{93.9} & \cellcolor{secondbest}\graybold{92.3} \\
    Comp & \cellcolor{best}\textbf{92.3} & \cellcolor{secondbest}\graybold{90.4} & 85.9 & 86.7 \\
    Qual & 78.2 & 78.7 & \cellcolor{best}\textbf{81.3} & \cellcolor{secondbest}\graybold{81.1} \\
    F1 & 13.6 & 14.1 & \cellcolor{secondbest}\graybold{14.8} & \cellcolor{best}\textbf{15.4} \\
    APLS & 59.65 & 67.53 & \cellcolor{best}\textbf{78.81} & \cellcolor{secondbest}\graybold{78.04} \\
    \hline
    rAULC & 19.56 & \cellcolor{secondbest}\graybold{78.65} & 69.03 & \cellcolor{best}\textbf{79.27} \\
    Corr (unc.) & 32.50 & \cellcolor{secondbest}\graybold{76.39} & 76.51 & \cellcolor{best}\textbf{87.46} \\
    ECE & \cellcolor{secondbest}\graybold{0.558} & 0.794 & 0.746 & \cellcolor{best}\textbf{0.419} \\
    ROC-AUC & 61.25 & 67.03 &\cellcolor{best} \textbf{68.42} & \cellcolor{secondbest} \graybold{67.09} \\
    PR-AUC & 62.64 & 67.85 &\cellcolor{secondbest} \graybold{71.25} &\cellcolor{best}\textbf{72.11} \\
    \hline
    Train & \cellcolor{best}\textbf{1x} & 5x & 14x & \cellcolor{secondbest}\graybold{2.8x} \\
    Inf & \cellcolor{secondbest}\graybold{5x} & \cellcolor{secondbest}\graybold{5x} & 13.5x & \cellcolor{best}\textbf{2.7x} \\
    \end{tabular}
    \caption{\small {\bf Uncertainty and performance metrics on MS Dataset.} The best result in each category is in \textbf{bold} and the second best is in \graybold{bold}. Most correspond to Ours and Iterative Ensembles.}
    \label{tab:ms_dataset_results}
\end{table}

\end{document}